\documentclass{article}

\PassOptionsToPackage{numbers, compress}{natbib}

\usepackage[final]{neurips_2020}

\usepackage[utf8]{inputenc} 
\usepackage[T1]{fontenc}    
\usepackage{hyperref}       
\usepackage{url}            
\usepackage{booktabs}       
\usepackage{amsfonts}       
\usepackage{nicefrac}       
\usepackage{microtype}      
\usepackage{amsmath}
\usepackage{algorithm}
\usepackage{algpseudocode}
\usepackage[pdftex]{graphicx}
\usepackage{wrapfig}
\usepackage{siunitx}
\usepackage{subcaption}
\usepackage{flexisym}
\usepackage[font=small,labelfont=bf]{caption}

\title{La-MAML: Look-ahead Meta Learning for Continual Learning}

\author{%
  Gunshi Gupta \thanks{equal contribution}\\
  Mila, UdeM \\
  \texttt{guptagun@mila.quebec} 
  \And
  Karmesh Yadav {\scriptsize *} \\
  Carnegie Mellon University\\
  \texttt{karmeshy@andrew.cmu.edu}
  \And
  Liam Paull \\
  Mila, UdeM \\
  \texttt{paulll@iro.umontreal.ca}
}

\begin{document}
\bibliographystyle{plainnat}
\maketitle

\begin{abstract}
The continual learning problem involves training models with limited capacity to perform well on a set of an unknown number of sequentially arriving tasks. 
While meta-learning shows great potential for reducing interference between old and new tasks, the current training procedures tend to be either slow or offline, and sensitive to many hyper-parameters. In this work, we propose \emph{Look-ahead MAML (La-MAML)}, a fast optimisation-based meta-learning algorithm for \emph{online}-continual learning, aided by a small episodic memory. Our proposed modulation of per-parameter learning rates in our meta-learning update allows us to draw connections to prior work on \emph{hypergradients} and \emph{meta-descent}. This provides a more flexible and efficient way to mitigate \emph{catastrophic forgetting} compared to conventional \emph{prior-based} methods.
\emph{La-MAML} achieves performance superior to other replay-based, prior-based and meta-learning based approaches for continual learning on real-world visual classification benchmarks.

\end{abstract}

\section{Introduction}

Embodied or interactive agents that accumulate knowledge and skills over time must possess the ability to continually learn.
\emph{Catastrophic forgetting} \cite{catastrophic, catastrophic2}, one of the biggest challenges in this setup, can occur when the \emph{i.i.d.} sampling conditions required by stochastic gradient descent (SGD) are violated as the data belonging to different tasks to be learnt arrives sequentially. Algorithms for \emph{continual learning} (CL) must also use their limited model capacity efficiently since the number of future tasks is unknown. 
Ensuring gradient-alignment across tasks is therefore essential, to make shared progress on their objectives. \emph{Gradient Episodic Memory} (GEM) \cite{lopez2017gradient} investigated the connection between weight sharing and forgetting in CL and developed an algorithm that explicitly tried to minimise \emph{gradient interference}. This is an objective that meta-learning algorithms implicitly optimise for (refer to \cite{nichol2018first} for derivations of the effective parameter update made in first and second order meta learning algorithms). \emph{Meta Experience Replay} (MER) \cite{riemer2018learning} formalized the transfer-interference trade-off and showed that the gradient alignment objective of GEM coincide with the objective optimised by the first order meta-learning algorithm Reptile \cite{nichol2018first}. 

Besides aligning gradients, meta-learning algorithms show promise for CL since they can \emph{directly} use the meta-objective to influence model optimisation and improve on auxiliary objectives like generalisation or transfer. This avoids having to define heuristic incentives like sparsity \cite{2018sr-nn} for better CL. The downside is that they are usually slow and hard to tune, effectively rendering them more suitable for \emph{offline} continual learning \cite{javed2019meta, riemer2018learning}.
In this work, we overcome these difficulties and develop a gradient-based meta-learning algorithm for \emph{efficient, online} continual learning. 
We first propose a base algorithm for continual meta-learning referred to as Continual-MAML (C-MAML) that utilizes a replay-buffer and optimizes a meta-objective that mitigates forgetting. Subsequently, we propose a modification to C-MAML, named La-MAML, which incorporates modulation of per-parameter learning rates (LRs) to pace the learning of a model across tasks and time. Finally, we show that the algorithm is scalable, robust  and achieves favourable performance on several benchmarks of varying complexity. 


\section{Related work}
Relevant CL approaches can be roughly categorized into \emph{replay-based, regularisation (or prior-based)} and \emph{meta-learning-based} approaches.

In order to  circumvent the issue of catastrophic forgetting, \emph{replay-based methods} maintain a collection of samples from previous tasks in memory. Approaches utilising an \emph{episodic-buffer} \cite{castro2018end, rebuffi2017icarl} uniformly sample old data points to mimic the \emph{i.i.d.} setup within continual learning. \emph{Generative-replay} \cite{2017genreplay} trains generative models to be able to replay past samples, with scalability concerns arising from the difficulty of modeling complex non-stationary distributions. GEM \cite{lopez2017gradient} and A-GEM \cite{chaudhry2018agem} take memory samples into account to determine altered \emph{low-interference} gradients for updating parameters. 

\emph{Regularisation-based} methods avoid using replay at all by constraining the network weights according to heuristics intended to ensure that performance on previous tasks is preserved. This involves penalising changes to weights deemed important for old tasks \cite{Kirkpatrick3521} or enforcing weight or representational sparsity \cite{2018selflesseq} to ensure that only a subset of neurons remain active at any point of time. The latter method has been shown to reduce the possibility of catastrophic interference across tasks \cite{2018sr-nn, serra2018hat}. 

\emph{Meta-Learning-based} approaches are fairly recent and have shown impressive results on small benchmarks like Omniglot and MNIST.
MER \cite{riemer2018learning}, inspired by GEM\cite{lopez2017gradient}, utilises replay to incentivise alignment of gradients between old and new tasks. Online-aware Meta Learning (OML) \cite{javed2019meta} introduces a meta-objective for a pre-training algorithm to learn an optimal representation \emph{offline}, which is subsequently frozen and used for CL.
\cite{2017online-meta-learn, 2019finn-onlineml, 2018onlinelearning} investigate orthogonal setups in which a learning agent uses all previously seen data to adapt quickly to an incoming stream of data, thereby ignoring the problem of catastrophic forgetting.
Our motivation lies in developing a \emph{scalable}, \emph{online} algorithm capable of learning from limited cycles through streaming data with reduced interference on old samples. In the following sections, we review background concepts and outline our proposed algorithm. We also note interesting connections to prior work not directly pertaining to CL.

\section{Preliminaries}
We consider a setting where a sequence of $T$ tasks $[\tau_1, \tau_2, .. \tau_T]$ is learnt by observing their training data [$D_1, D_2, .. D_T$] sequentially. 
We define ${X^i,Y^i}$ = $\{{(x^i_n, y^i_n)}\}_{n=0}^{N_{i}}$ as the set of $N_{i}$ input-label pairs randomly drawn from $D_i$. 
An any time-step $j$ during online learning, we aim to minimize the empirical risk of the model on all the $t$ tasks seen so far ($\tau_{1:t}$), given limited access to data $(X^i,Y^i)$ from previous tasks $\tau_i$ ($i < t$).
We refer to this objective as the \emph{cumulative risk}, given by:
\begin{equation}
        \sum_{i=1}^{{t}} \mathbb{E}_{\left({X}^{i}, {Y}^{i}\right)}\left[\ell_{i} \left(f_{i}\left({X}^{i} ; \theta\right), {Y}^{i}\right)\right] 
= 
        \mathbb{E}_{\left({X}^{1:t}, {Y}^{1:t}\right)}\left[L_{t}\left(f\left({X}^{1:t} ; \theta\right), {Y}^{1:t}\right)\right]
\label{eq:CL}
\end{equation}
where $\ell_{i}$ is the loss on $\tau_i$ and $f_{i}$ is a learnt, possibly task-specific mapping from inputs to outputs using parameters $\theta^j_0$. $L_{t} = \sum_{i=1}^{t} \ell_{i}$ is the sum of all task-wise losses for tasks $\tau_{1:t}$ where $t$ goes from $1$ to $T$. 
Let $\ell$ denote some loss objective to be minimised. Then the SGD operator acting on parameters $\theta^j_0$, denoted by $U(\theta^j_0)$ is defined as: 
\begin{align}
   U \left(\theta^{j}_0\right) = \theta^{j}_{1} =\theta^{j}_0-\alpha \nabla_{\theta^{j}_0} \ell(\theta^{j}_0) =\theta^{j}_0-\alpha g^{j}_0
\end{align}
where  $g^{j}_0 = \nabla_{\theta^{j}_0} \ell(\theta^{j}_0)$.
$U$ can be composed for $k$ updates as $U_k \left(\theta^{j}_0\right)= U ... \circ U \circ U(\theta^{j}_0) = \theta^{j}_k$. $\alpha$ is a scalar or a vector LR. $U\left(\cdot, x\right)$ implies gradient updates are made on data sample $x$.
We now introduce the MAML \cite{finn2017model} and OML \cite{javed2019meta} algorithms, that we build upon in Section \ref{proposed}. 

\textbf{Model-Agnostic Meta-Learning (MAML)}:
Meta-learning \cite{Schmidhuber1987}, or \emph{learning-to-learn} \cite{learn2learn} has emerged as a popular approach for training models amenable to fast adaptation on limited data. 
MAML \cite{finn2017model} proposed optimising model parameters to learn a set of tasks \emph{while} improving on auxiliary objectives like few-shot generalisation within the task distributions. 
We review some common terminology used in gradient-based meta-learning: 
1) at a given time-step $j$ during training, model parameters $\theta^{j}_0$ (or $\theta_0$ for simplicity), are often referred to as an \emph{initialisation}, since the aim is to find an ideal starting point for few-shot gradient-based adaptation on unseen data.
2) \emph{Fast} or \emph{inner-updates}, refer to gradient-based updates made to a copy of $\theta_0$, optimising some inner objective (in this case, $\ell_{i}$ for some $\tau_i$).
3) A \emph{meta-update} involves the \emph{trajectory} of fast updates from $\theta_0$ to $\theta_k$, followed by making a permanent gradient update (or \emph{slow-update}) to $\theta_0$. This \emph{slow-update} is computed by evaluating an auxiliary objective (or \emph{meta-loss} $L_{meta})$ on $\theta_k$, and differentiating through the \emph{trajectory} to obtain $\nabla_{\theta_0} L_{meta}(\theta_k)$.
MAML thus optimises $\theta^j_0$ at time $j$, to perform optimally on tasks in $\{\tau_{1:t}\}$ after undergoing a few gradient updates on their samples. It optimises in every \emph{meta-update}, the objective:
\begin{equation}
    \min_{\theta_0^j} \mathbb{E}_{\tau_{1:t}}\left[L_{meta}\left(U_{k}(\theta_0^j)\right)\right]
    = \min_{\theta^j_0} \mathbb{E}_{\tau_{1:t}}\left[L_{meta}(\theta_k^j)\right].
\end{equation}

\textbf{Equivalence of Meta-Learning and CL Objectives}: The approximate equivalence of first and second-order meta-learning algorithms like Reptile and MAML was shown in \cite{nichol2018first}. 
MER \cite{riemer2018learning} then showed that their CL objective of minimising loss on and aligning gradients between a set of tasks $\tau_{1:t}$ seen till any time $j$ \emph{(on the left)}, can be optimised by the Reptile objective \emph{(on the right)}, ie. : \\
\begin{equation}
    \min _{\theta^j_0} \left( \sum_{i=1}^{t} \left(\ell_{i}(\theta_0^j)\right) - \alpha \sum_{p,q \leq t} \left( \frac{\partial \ell_{p}\left(\theta^j_0\right)}{\partial \theta^j_0} \cdot \frac{\partial \ell_{q}\left(\theta^j_0\right)}{\partial \theta^j_0} \right) \right) = \min _{\theta^j_0}   \mathbb{E}_{\tau_{1:t}}\left[L_{t}\left(U_{k}(\theta^j_0)\right)\right]
    \label{mer-objective}
\end{equation}
where the \emph{meta-loss} $L_{t} = \sum_{i=1}^{t} \ell_{i}$ is evaluated on samples from tasks $\tau_{1:t}$. This implies that the procedure to meta-learn an \emph{initialisation} coincides with learning optimal parameters for CL.

\textbf{Online-aware Meta-Learning (OML)}: \cite{javed2019meta} proposed to meta-learn a \emph{Representation-Learning Network (RLN)} to provide a representation suitable for CL to a \emph{Task-Learning Network (TLN)}. The RLN's representation is learnt in an \emph{offline} phase, where it is trained using \emph{catastrophic forgetting as the learning signal}. Data from a fixed set of tasks ($\tau_{val}$), is repeatedly used to evaluate the RLN and TLN as the TLN undergoes temporally correlated updates. In every \emph{meta-update}'s inner loop,  the TLN undergoes \emph{fast updates} on streaming task data with a frozen RLN. The RLN and updated TLN are then evaluated through a \emph{meta-loss} computed on data from $\tau_{val}$ along with the current task. This tests how the performance of the model has changed on $\tau_{val}$ in the process of trying to learn the streaming task. The meta-loss is then differentiated to get gradients for \emph{slow updates} to the TLN and RLN.
This composition of two losses to simulate CL in the inner loop and test \emph{forgetting} in the outer loop, is referred to as the \emph{OML objective}. The RLN learns to eventually provide a better representation to the TLN for CL, one which is shown to have emergent sparsity. 

\section{Proposed approach}
\label{proposed}
In the previous section, we saw that the OML objective can directly regulate CL behaviour, and that MER exploits the approximate equivalence of meta-learning and CL objectives. We noted that OML trains a static representation \emph{offline} and that MER's algorithm is prohibitively slow. We show that optimising the OML objective \emph{online} through a multi-step MAML procedure is equivalent to a more sample-efficient CL objective. 
In this section, we describe \emph{Continual-MAML} (C-MAML), the base algorithm that we propose for online continual learning. 
We then detail an extension to C-MAML, referred to as Look-Ahead MAML (La-MAML), outlined in Algorithm \ref{alg:alphamaml}.

\subsection{C-MAML}
C-MAML aims to optimise the OML objective \emph{online}, so that learning on the current task doesn't lead to forgetting on previously seen tasks.
We define this objective, adapted to optimise a model's parameters $\theta$ instead of a representation at time-step $j$, as:
\begin{equation}
    \min _{\theta_0^j} \operatorname{OML}(\theta_0^j, t)
      =  \min _{\theta_0^j} \sum_{\mathcal{S}_{k}^{j} \sim D_{t}}\left[L_{t}\left(U_k(\theta^j_0, \mathcal{S}_{k}^{j})\right)\right]
  \label{eq:1}
\end{equation}

where $S_{k}^{j}$ is a stream of $k$ data tuples $\left(X_{j+l}^{t}, Y_{j+l}^{t}\right)_{l=1}^{k}$ from the current task $\tau_t$ that is seen by the model at time $j$. The meta-loss $L_{t} = \sum_{i=1}^{t} \ell_{i}$ is evaluated on $\theta^j_k = U_k(\theta^j_0, S_{k}^{j})$. It evaluates the fitness of $\theta^j_k$ for the continual learning prediction task defined in Eq.~\ref{eq:CL} until $\tau_t$. 
We omit the implied data argument $(x^i, y^i)\sim(X^{i},Y^{i})$ that is the input to each loss $\ell_{i}$ in $L_t$ for any task $\tau_i$.
We will show in Appendix \ref{objective} that optimising our objective in Eq. \ref{eq:1} through the $k$-step MAML update in C-MAML also coincides with optimising the CL objective of AGEM \cite{chaudhry2018agem}:
\begin{equation}
    \min _{\theta^j_0} \mathbb{E}_{\tau_{1:t}}\left[L_{t}\left(U_{k}(\theta^j_0)\right)\right] = \min _{\theta^j_0} \sum_{i=1}^{t} \left( \ell_{i}(\theta^j_0) - \alpha \frac{\partial \ell_{i}\left(\theta^j_0\right)}{\partial \theta^j_0} \cdot \frac{\partial \ell_{t}\left(\theta^j_0\right)}{\partial \theta^j_0} \right).
    \label{our-objective}
\end{equation}
This differs from Eq.  \ref{mer-objective}'s objective by being \emph{asymmetric}: it focuses on aligning the gradients of $\tau_{t}$ and the average gradient of $\tau_{1:t}$ instead of aligning all the pair-wise gradients between tasks $\tau_{1:t}$. 
In Appendix \ref{representation}, we show empirically that gradient alignment amongst old tasks doesn't degrade while a new task is learnt, avoiding the need to repeatedly optimise the inter-task alignment between them.
This results in a drastic speedup over MER's objective (Eq. \ref{mer-objective}) which tries to align all $\tau_{1:t}$ equally, thus resampling incoming samples $s\sim{\tau_t}$ to form a uniformly distributed batch over $\tau_{1:t}$. Since each $s$ then has $\frac{1}{t}$-th the contribution in gradient updates, it becomes necessary for MER to take multiple passes over many such uniform batches including $s$. 

\begin{wrapfigure}{R}{.5\textwidth}
    \begin{minipage}{1.0\linewidth}
    \begin{figure}[H]
        \begin{center}
            \centerline{\includegraphics[width=1.0\linewidth]{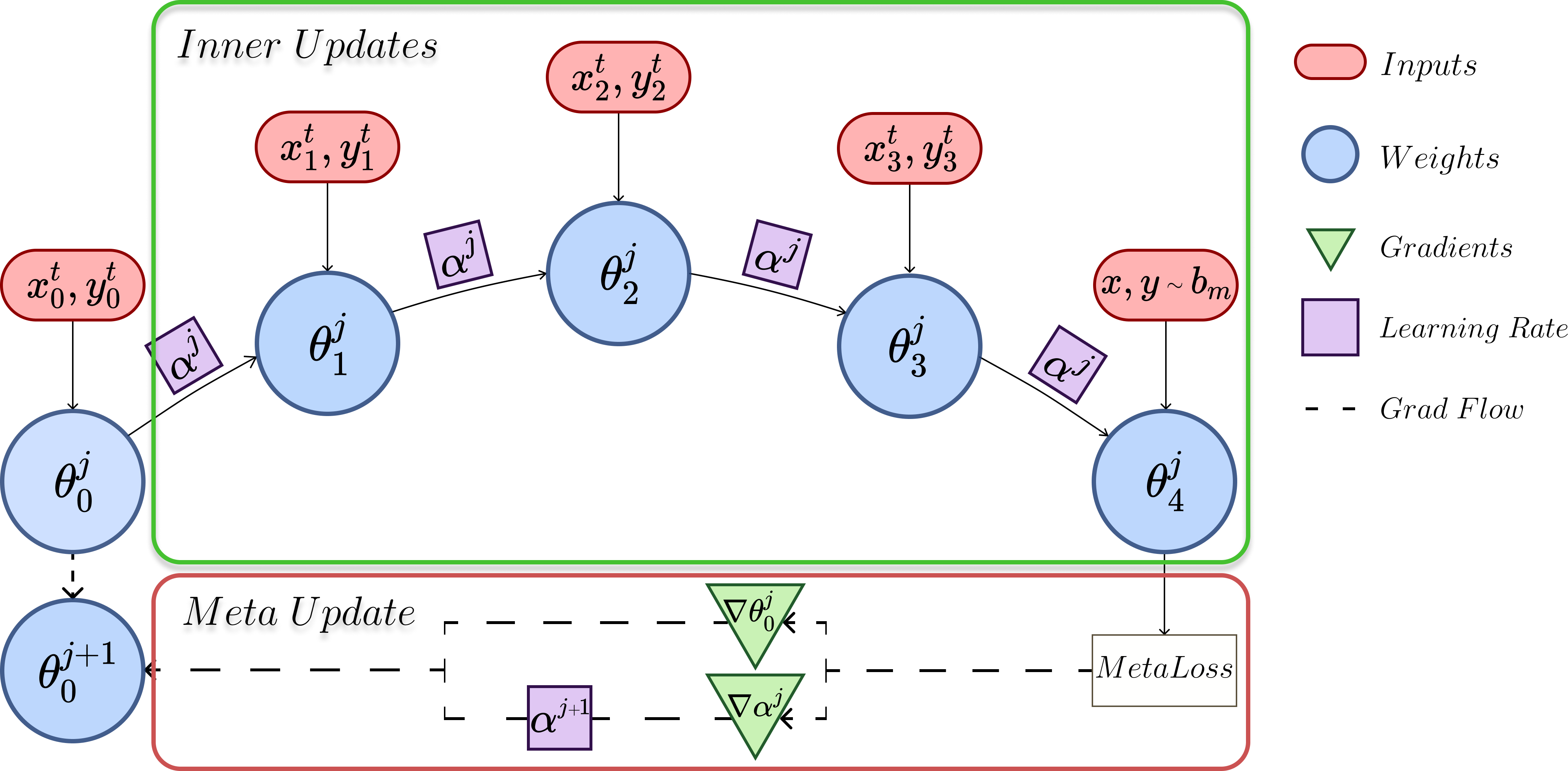}}
            \caption{The proposed \textbf{La-MAML} algorithm: For every batch of data, the initial weights undergo a series of $k$ \emph{fast updates} to obtain $\theta_k^j$ (here $j=0$), which is evaluated against a meta-loss to backpropagate gradients with respect to the weights $\theta_0^0$  and LRs $\alpha^0$. First $\alpha^0$ is updated to $\alpha^1$ which is then used to update $\theta_0^0$ to $\theta_0^1$  The blue boxes indicate \emph{fast weights} while the green boxes indicate gradients for the \emph{slow updates}. LRs and weights are updated in an asynchronous manner.
            \label{alpha-maml-figure}}
        \end{center}
        \vskip -0.5in
    \end{figure}
    \end{minipage}
\end{wrapfigure}
 

 During training, a replay-buffer $R$ is populated through \emph{reservoir sampling} on the incoming data stream as in \cite{riemer2018learning}.
At the start of every meta-update, a batch $b$ is sampled from the current task. $b$ is also combined with a batch sampled from $R$ to form the \emph{meta-batch}, $b_m$, representing samples from both old and new tasks. 
$\theta_0^j$ is updated through $k$ SGD-based \emph{inner-updates} by seeing the current task's samples from $b$ one at a time. 
The outer-loss or \emph{meta-loss} $L_{t}(\theta^j_k)$ is evaluated on $b_m$. 
It indicates the performance of parameters $\theta^j_k$ on all the tasks $\tau_{1:t}$ seen till time $j$. 
The complete training procedure is described in Appendix \ref{cmamlalgo}.

\subsection{La-MAML}
Despite the fact that meta-learning incentivises the alignment of \emph{within-task} and \emph{across-task} gradients, there can still be some interference between the gradients of old and new tasks, $\tau_{1:{t-1}}$ and $\tau_t$ respectively. This would lead to forgetting on $\tau_{1:t-1}$, since its data is no longer fully available to us. This is especially true at the beginning of training a new task, when its gradients aren't necessarily aligned with the old ones. A mechanism is thus needed to ensure that \emph{meta-updates} are conservative with respect to $\tau_{1:t-1}$, so as to avoid negative transfer on them. The magnitude and direction of the \emph{meta-update} needs to be regulated, guided by how the loss on $\tau_{1:t-1}$ would be affected by the update.

We propose \textbf{Lookahead-MAML (La-MAML)}, where we include a set of learnable per-parameter learning rates (LRs) to be used in the \emph{inner updates}, as depicted in Figure \ref{alpha-maml-figure}. 
This is motivated by our observation that the expression for the gradient of Eq. \ref{eq:1} with respect to the inner loop's LRs directly reflects the alignment between the old and new tasks.
The augmented learning objective is defined as
\begin{equation}
   \min _{\theta^j_0, \alpha^j} \sum_{\mathcal{S}_{k}^{j} \sim D_t}\left[L_{t}\left(U_k\left(\alpha^j, \theta_0^j, \mathcal{S}_{k}^{j}\right)\right)\right],
\label{eq:alpha-maml-objective}
\end{equation}

and the gradient of this objective at time $j$, with respect to the LR vector $\alpha^j$ (denoted as  $g_{MAML}(\alpha^j)$) is then given as: 
   
\begin{equation}
     g_{MAML}(\alpha^j) = \frac{\partial}{\partial \alpha^j} L_{t}\left(\theta^j_{k}\right) = 
     \frac{\partial}{\partial \theta^j_{k}} L_{t}\left(\theta^j_{k}\right) \cdot \left( -  \sum_{k^{\prime}=0}^{k-1} \frac{\partial}{\partial \theta^j_{k^{\prime}}} \ell_{t} \left(\theta^j_{k^{\prime}}\right)  \right).
  \label{eq:alpha-maml}
\end{equation}

We provide the full derivation in the Appendix \ref{hypergrad}, and simply state the expression for a first-order approximation \cite{finn2017model} of $g_{MAML}(\alpha)$ here. The first term in $g_{MAML}(\alpha)$ corresponds to the gradient of the meta-loss on batch $b_m$: $g_{meta}$. The second term indicates the cumulative gradient from the inner-updates: $g_{traj}$. 
This expression indicates that the gradient of the LRs will be negative when the inner product between $g_{meta}$ and $g_{traj}$ is high, ie. the two are aligned; zero when the two are orthogonal (not interfering) and positive when there is interference between the two. Negative (positive) LR gradients would pull up (down) the LR magnitude. We depict this visually in Figure \ref{fig:gradients}.

\begin{minipage}{\textwidth}
    \begin{minipage}[b]{0.64\linewidth}
        \begin{algorithm}[H] 
           \caption{La-MAML : Look-ahead MAML}
           \label{alg:alphamaml}
        \begin{algorithmic}
           \State {\bfseries Input:} Network weights $\theta$, LRs $\alpha$, inner objective $\ell$, meta objective $L$, learning rate for $\alpha$ : $\eta$  
           \State $j \leftarrow 0$, $R \leftarrow \{ \}$ \Comment{Initialise Replay Buffer} 
        
           \For{$t:=1$ {\bfseries to} $T$}
            \For{$ep:=1$ {\bfseries to} $num_{epochs}$}
           
                \For{batch $b$ {\bfseries in} $(X^t,Y^t)\sim D_t$}
                    \State $k \leftarrow sizeof(b)$
                    \State $b_m \leftarrow Sample(R) \cup b$
                   
                    \For{$n=0$ {\bfseries to} $k-1$}
                        \State Push $b[k^{\prime}]$ to R with reservoir sampling         
                        \State $\theta_{k^{\prime}+1}^{j} \leftarrow  \theta_{k^{\prime}}^{j}  - \alpha^{j} \cdot \nabla_{\theta_{k^{\prime}}^{j}}$
                    \EndFor
                    \State $\alpha^{j+1} \leftarrow \alpha^{j} - \eta \nabla_{\alpha^{j}} L_{t}(\theta_k^j, b_m)$  \hfill (a)
                    \State $\theta^{j+1}_{0} \leftarrow \theta_0^{j} - max(0, \alpha^{j+1}) \cdot \nabla_{\theta_0^{j}}  L_{t}(\theta_k^j, b_m)$ \hfill (b)
                    \State $j \leftarrow j+1$
                \EndFor
            \EndFor
            \EndFor
        \end{algorithmic}
        \end{algorithm}
    \end{minipage}
      \hfill
    \begin{minipage}[b]{0.35\linewidth}
    \begin{figure}[H] 
        \begin{center}
        \centerline{\includegraphics[width=1.0\linewidth]{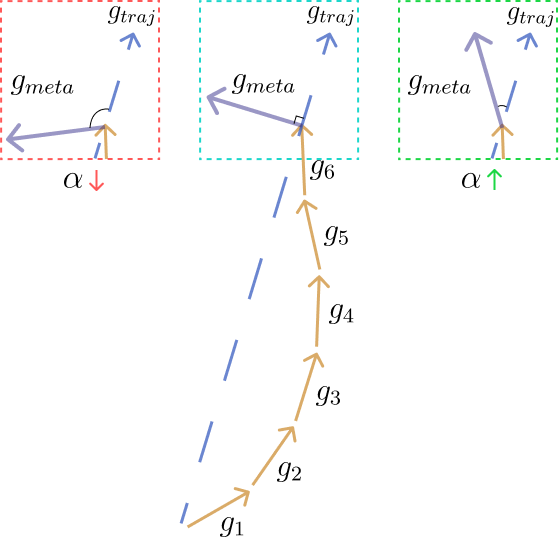}}
        \caption{Different scenarios for the alignment of $g_{traj}$ (blue dashed line) and $g_{meta}$, going from interference (left) to alignment (right). Yellow arrows denote the \emph{inner updates}. The LR $\alpha$ increases (decreases) when gradients align (interfere).}
        \label{fig:gradients}
        \end{center}
        \vskip -0.2in
    \end{figure}
    \end{minipage}
\end{minipage}

We propose updating the network weights and LRs \emph{asynchronously} in the meta-update.
Let $\alpha^{j+1}$ be the updated LR vector obtained by taking an SGD step with the LR gradient from Eq. \ref{eq:alpha-maml} at time $j$. We then update the weights as:
\begin{equation}
     \theta_0^{j+1} \leftarrow \theta_0^j - max(0, \alpha^{j+1}) \cdot \nabla_{\theta^j_0} L_{t}(\theta^j_k)  
\end{equation}
where $k$ is the number of steps taken in the inner-loop. The LRs $\alpha^{j+1}$ are clipped to positive values to avoid \emph{ascending} the gradient, and also to avoid making \emph{interfering} parameter-updates, thus mitigating catastrophic forgetting. 
The meta-objective thus conservatively modulates the pace and direction of learning to achieve quicker learning progress on a new task while facilitating transfer on old tasks. Algorithm \ref{alg:alphamaml} \footnote{The code for our algorithm can be found at: \url{https://github.com/montrealrobotics/La-MAML}} illustrates this procedure.
Lines (a), (b) are the only difference between C-MAML and La-MAML, with C-MAML using a fixed scalar LR $\alpha$ for the meta-update to $\theta^j_0$ instead of $\alpha^{j+1}$.


Our meta-learning based algorithm incorporates concepts from both prior-based and replay-based approaches. The LRs modulate the parameter updates in an data-driven manner, guided by the interplay between gradients on the replay samples and the streaming task.
However, since LRs evolve with every meta-update, their decay is temporary. This is unlike many prior-based approaches, where penalties on the change in parameters gradually become so high that the network capacity saturates \cite{Kirkpatrick3521}. Learnable LRs can be modulated to high and low values as tasks arrive, thus being a simpler, flexible and elegant way to constrain weights. 
This asynchronous update resembles trust-region optimisation \cite{trustregion} or \emph{look-ahead search} since the step-sizes for each parameter are adjusted based on the loss incurred after applying hypothetical updates to them.
Our LR update is also analogous to the heuristic uncertainty-based LR update in UCB \cite{ebrahimi2019uncertainty}, BGD \cite{bgd}, which we compare to in Section \ref{expt-modulation}.



\subsection{Connections to Other Work}
\label{other-work}
\textbf{Stochastic Meta-Descent (SMD)}: 
When learning over a non-stationary data distribution, using decaying LR schedules is not common. Strictly diminishing LR schedules aim for closer \emph{convergence} to a fixed mimima of a stationary distribution, which is at odds with the goal of online learning. 
It is also not possible to manually tune these schedules since the extent of the data distribution is unknown. 
However, \emph{adaptivity} in LRs is still highly desired to adapt to the optimisation landscape, accelerate learning and modulate the degree of adaptation to reduce catastrophic forgetting. Our adaptive LRs can be connected to work on \emph{meta-descent} \cite{hypergradient, smd} in \emph{offline} supervised learning (OSL). 
While several variations of \emph{meta-descent} exist, the core idea behind them and our approach is \emph{gain adaptation}.
While we adapt the gain based on the correlation between old and new task gradients to make shared progress on all tasks, \cite{hypergradient, smd} use the correlation between two successive stochastic gradients to converge faster.
We rely on the meta-objective's differentiability with respect to the LRs, to obtain LR \emph{hypergradients} automatically.

\textbf{Learning LRs in meta-learning}: Meta-SGD \cite{li2017metasgd} proposed learning the LRs in MAML for few-shot learning. Some notable differences between their update and ours exist. They \emph{synchronously} update the weights and LRs while our \emph{asynchronous} update to the LRs serves to carry out a more conservative update to the weights. The intuition for our update stems from the need to mitigate gradient interference and its connection to the transfer-interference trade-off ubiquitous in continual learning.
$\alpha$-MAML \cite{alphamaml} analytically updates the two \emph{scalar} LRs in the MAML update for more adaptive few-shot learning. Our \emph{per-parameter} LRs are modulated implicitly through back-propagation, to regulate change in parameters based on their alignment across tasks, providing our model with a more powerful degree of adaptability in the CL domain.

\section{Experiments}
\label{experiments}
In this section, we evaluate La-MAML in settings where the model has to learn a set of sequentially streaming classification tasks.  
\emph{Task-agnostic} experiments, where the task identity is unknown at training and test-time, are performed on the MNIST benchmarks with a \emph{single-headed} model. 
\emph{Task-aware} experiments with known task identity, are performed on the CIFAR and TinyImagenet \cite{tinyimagenet} datasets with a \emph{multi-headed} model.
Similar to \cite{riemer2018learning}, we use the retained accuracy (RA) metric to compare various approaches. RA is the average accuracy of the model across tasks at the end of training. We also report the \emph{backward-transfer and interference} (BTI) values which measure the average change in the accuracy of each task from when it was learnt to the end of the last task. A smaller BTI implies lesser forgetting during training.
\begin{table*}[b]\centering
\caption{RA, BTI and their standard deviation on MNIST benchmarks. Each experiment is run with 5 seeds.}
\label{acc-table}
\begin{center}
\begin{small}
\begin{sc}
\begin{tabular}{@{\extracolsep{2pt}}lcccccc@{}}
\toprule
Method      &                   \multicolumn{2}{c}{Rotations}          &                \multicolumn{2}{c}{Permutations}          &                 \multicolumn{2}{c}{Many}                \\
\cmidrule{2-3}  \cmidrule{4-5}  \cmidrule{6-7}  
            &               RA           &              BTI            &             RA             &            BTI              &             RA             &            BTI             \\
\midrule
Online      & 53.38 {\tiny $\pm$ 1.53} &  -5.44 {\tiny $\pm$ 1.70} & 55.42 {\tiny $\pm$ 0.65} & -13.76 {\tiny $\pm$ 1.19} & 32.62 {\tiny $\pm$ 0.43} & -19.06 {\tiny $\pm$ 0.86} \\
EWC         & 57.96 {\tiny $\pm$ 1.33} & -20.42 {\tiny $\pm$ 1.60} & 62.32 {\tiny $\pm$ 1.34} & -13.32 {\tiny $\pm$ 2.24} & 33.46 {\tiny $\pm$ 0.46} & -17.84 {\tiny $\pm$ 1.15} \\
GEM         & 67.38 {\tiny $\pm$ 1.75} & -18.02 {\tiny $\pm$ 1.99} & 55.42 {\tiny $\pm$ 1.10} & -24.42 {\tiny $\pm$ 1.10} & 32.14 {\tiny $\pm$ 0.50} & -23.52 {\tiny $\pm$ 0.87} \\
MER         & \textbf{77.42 {\tiny $\pm$ 0.78}} &\textbf{-5.60{\tiny$\pm$0.70}} & 73.46{\tiny $\pm$ 0.45} &  -9.96 {\tiny $\pm$ 0.45} & 47.40 {\tiny $\pm$ 0.35} & -17.78 {\tiny $\pm$ 0.39} \\
\midrule
C-MAML      & 77.33 {\tiny $\pm$ 0.29} &  -7.88 {\tiny $\pm$ 0.05} & \textbf{74.54 {\tiny $\pm$ 0.54}} & -10.36 {\tiny $\pm$ 0.14} & 47.29 {\tiny $\pm$ 1.21} & -20.86 {\tiny $\pm$ 0.95} \\
Sync        & 74.07 {\tiny $\pm$ 0.58} & -6.66 {\tiny $\pm$ 0.44}  & 70.54 {\tiny $\pm$ 1.54} & -14.02 {\tiny $\pm$ 2.14} & 44.48 {\tiny $\pm$ 0.76} & -24.18 {\tiny $\pm$ 0.65} \\
La-MAML     & \textbf{77.42 {\tiny $\pm$ 0.65}} & -8.64 {\tiny $\pm$ 0.403} & 74.34 {\tiny $\pm$ 0.67} & \textbf{-7.60 {\tiny $\pm$ 0.51}} & \textbf{48.46 {\tiny $\pm$ 0.45}} & \textbf{-12.96 {\tiny $\pm$ 0.073}} \\ 
\bottomrule
\end{tabular}
\end{sc}
\end{small}
\end{center}
\end{table*}

\emph{Efficient Lifelong Learning (LLL)}: 
Formalized in \cite{chaudhry2018agem}, the setup of efficient lifelong learning assumes that incoming data for every task has to be processed in only one single pass: once processed, data samples are not accessible anymore unless they were added to a replay memory. We evaluate our algorithm on this challenging \emph{(Single-Pass)} setup as well as the standard \emph{(Multiple-Pass)} setup, where ideally offline training-until-convergence is performed for every task, once we have access to the data.

\begin{wrapfigure}{R}{.5\textwidth}
    \begin{minipage}{1.0\linewidth}
        \vskip -0.2in
        \begin{table}[H]
            \caption{Running times for MER and La-MAML on MNIST benchmarks for one epoch}
            \label{timing-table}
            \begin{center}
            \begin{small}
            \begin{sc}
            \begin{tabular}{lcc}
            \toprule
            Method & Rotations & Permutations \\
            \midrule
            La-MAML &  45.95 {\tiny $\pm$ 0.38} &  46.13 {\tiny $\pm$ 0.42} \\
            MER  & 218.03 {\tiny $\pm$ 6.44} & 227.11 {\tiny $\pm$ 12.12}\\
            \bottomrule
            \end{tabular}
            \end{sc}
            \end{small}
            \end{center}
            \vskip -0.1in
        \end{table}
    \end{minipage}
\end{wrapfigure}

\subsection{Continual learning benchmarks}
First, we carry out experiments on the toy continual learning benchmarks proposed in prior CL works. \textbf{MNIST Rotations}, introduced in \cite{lopez2017gradient}, comprises tasks to classify MNIST digits rotated by a different common angle in [0, 180] degrees in each task. In \textbf{MNIST Permutations}, tasks are generated by shuffling the image pixels by a fixed random permutation. Unlike Rotations, the input distribution of each task is unrelated here, leading to less positive transfer between tasks. Both MNIST Permutation and MNIST Rotation have 20 tasks with 1000 samples per task. \textbf{Many Permutations}, a more complex version of Permutations, has five times more tasks (100 tasks) and five times less training data (200 images per task). Experiments are conducted in the low data regime with only 200 samples for Rotation and Permutation and 500 samples for Many, which allows the differences between the various algorithm to become prominent (detailed in Appendix \ref{experimental}). We use the same architecture and experimental settings as in MER \cite{riemer2018learning}, allowing us to compare directly with their results. We use the cross-entropy loss as the \emph{inner} and \emph{outer} objectives during meta-training.
Similar to \cite{nichol2018first}, we see improved performance when evaluating and summing the \emph{meta-loss} at all steps of the inner updates as opposed to just the last one.

We compare our method in the \emph{Single-Pass} setup against multiple baselines including  \emph{Online}, \emph{Independent}, \emph{EWC} \cite{Kirkpatrick3521}, \emph{GEM} \cite{lopez2017gradient} and \emph{MER} \cite{riemer2018learning} (detailed in Appendix \ref{baselines}), as well as different ablations (discussed in Section \ref{expt-modulation}). In Table \ref{acc-table}, we see that La-MAML achieves comparable or better performance than the baselines on all benchmarks. Table \ref{timing-table} shows that La-MAML matches the performance of MER in less than 20\% of the training time, owing to its sample-efficient objective which allows it to make make more learning progress per iteration. This also allows us to scale it to real-world visual recognition problems as described next.

\subsection{Real-world classification}
While La-MAML fares well on the MNIST benchmarks, we are interested in understanding its capabilities on more complex visual classification benchmarks.
We conduct experiments on the \textbf{CIFAR-100} dataset in a task-incremental manner \cite{lopez2017gradient} where, 20 tasks comprising of disjoint \emph{5-way} classification problems are streamed. We also evaluate on the \textbf{TinyImagenet-200} dataset by partitioning its 200 classes into 40 \emph{5-way} classification tasks. Experiments are carried out in both the \emph{Single-Pass} and \emph{Multiple-Pass} settings, where in the latter we allow all CL approaches to train up to a maximum of 10 epochs. Each method is allowed a replay-buffer, containing upto 200 and 400 samples for CIFAR-100 and TinyImagenet respectively. We provide further details about the baselines in Appendix \ref{baselines} and  about the architectures, evaluation setup and hyper-parameters in Appendix \ref{experimental}. 

\begin{table}[b]
\caption{Results on the standard continual (Multiple) and LLL (Single) setups with CIFAR-100 and TinyImagenet-200. Experiments are run with 3 seeds. * indicates result omitted due to high instability.  } 
\label{imgnet-table}
\begin{center}
\begin{scriptsize}
\begin{sc}
\begin{tabular}{@{\extracolsep{-5pt}}lcccccccc@{}}
\toprule
Method          &                                       \multicolumn{4}{c}{CIFAR-100}                                        &                                   \multicolumn{4}{c}{TinyImagenet}                                          \\
                                 \cline { 2 - 5 }                                                                                                           \cline { 6 -9 }     
                &       \multicolumn{2}{c}{Multiple}                  &             \multicolumn{2}{c}{Single}              &           \multicolumn{2}{c}{Multiple}             &             \multicolumn{2}{c}{Single}              \\
                &           RA             &           BTI            &           RA             &           BTI            &           RA            &           BTI            &           RA             &           BTI            \\
\midrule
IID             & 85.60 {\tiny $\pm$ 0.40} &            -             &            -             &            -             & 77.1 {\tiny $\pm$ 1.06} &            -             &            -             &            -             \\
\hline
ER              & 59.70 {\tiny $\pm$ 0.75} &-16.50 {\tiny $\pm$ 1.05} & 47.88 {\tiny $\pm$ 0.73} &-12.46 {\tiny $\pm$ 0.83} & 48.23 {\tiny $\pm$ 1.51} &-19.86 {\tiny $\pm$ 0.70} & 39.38 {\tiny $\pm$ 0.38} &-14.33 {\tiny $\pm$ 0.89} \\
iCARL           & 60.47 {\tiny $\pm$ 1.09} &-15.10 {\tiny $\pm$ 1.04} & 53.55 {\tiny $\pm$ 1.69} & \textbf{-8.03 {\tiny $\pm$ 1.16}} & 54.77 {\tiny $\pm$ 0.32} & \textbf{-3.93 {\tiny $\pm$ 0.55}} & 45.79 {\tiny $\pm$ 1.49} & \textbf{-2.73 {\tiny $\pm$ 0.45}} \\
GEM             & 62.80 {\tiny $\pm$ 0.55} &-17.00 {\tiny $\pm$ 0.26} & 48.27 {\tiny $\pm$ 1.10} & -13.7 {\tiny $\pm$ 0.70} & 50.57 {\tiny $\pm$ 0.61} &-20.50 {\tiny $\pm$ 0.10} & 40.56 {\tiny $\pm$ 0.79} &-13.53 {\tiny $\pm$ 0.65} \\
AGEM            & 58.37 {\tiny $\pm$ 0.13} &-17.03 {\tiny $\pm$ 0.72} & 46.93 {\tiny $\pm$ 0.31} & -13.4 {\tiny $\pm$ 1.44} & 46.38 {\tiny $\pm$ 1.34} &-19.96 {\tiny $\pm$ 0.61} & 38.96 {\tiny $\pm$ 0.47} &-13.66 {\tiny $\pm$ 1.73} \\
MER             &             -            &             -            & 51.38 {\tiny $\pm$ 1.05} &-12.83 {\tiny $\pm$ 1.44} &             -            &             -            & 44.87 {\tiny $\pm$ 1.43} &-12.53 {\tiny $\pm$ 0.58} \\
\midrule
Meta-BGD        & 65.09 {\tiny $\pm$ 0.77} &-14.83 {\tiny $\pm$ 0.40} & 57.44 {\tiny $\pm$ 0.95} & -10.6 {\tiny $\pm$ 0.45} &             *            &             *            & 50.64 {\tiny $\pm$ 1.98} & -6.60 {\tiny $\pm$ 1.73} \\
C-MAML           & 65.44 {\tiny $\pm$ 0.99} &-13.96 {\tiny $\pm$ 0.86} & 55.57 {\tiny $\pm$ 0.94} & -9.49 {\tiny $\pm$ 0.45} & 61.93 {\tiny $\pm$ 1.55} &-11.53 {\tiny $\pm$ 1.11} & 48.77 {\tiny $\pm$ 1.26} & -7.6  {\tiny $\pm$ 0.52} \\
La-ER & 67.17 {\tiny $\pm$ 1.14} &-12.63 {\tiny $\pm$ 0.60} & 56.12 {\tiny $\pm$ 0.61} & -7.63 {\tiny $\pm$ 0.90}  & 54.76 {\tiny $\pm$ 1.94} &-15.43 {\tiny $\pm$ 1.36} & 44.75 {\tiny $\pm$ 1.96} & -10.93 {\tiny $\pm$ 1.32}  \\
Sync            & 67.06 {\tiny $\pm$ 0.62} &-13.66 {\tiny $\pm$ 0.50} & 58.99 {\tiny $\pm$ 1.40} & -8.76 {\tiny $\pm$ 0.95} & 65.40 {\tiny $\pm$ 1.40} &-11.93 {\tiny $\pm$ 0.55} & \textbf{52.84 {\tiny $\pm$ 2.55}} & -7.3{\tiny $\pm$ 1.93} \\
La-MAML          & \textbf{70.08 {\tiny $\pm$ 0.66}} & \textbf{-9.36 {\tiny $\pm$ 0.47}} & \textbf{61.18 {\tiny $\pm$ 1.44}} & -9.00 {\tiny $\pm$ 0.2} & \textbf{66.99 {\tiny $\pm$ 1.65}} & \textbf{-9.13 {\tiny $\pm$ 0.90}} & 52.59 {\tiny $\pm$ 1.35} &  \textbf{-3.7 {\tiny $\pm$ 1.22}} \\

\bottomrule
\end{tabular}
\end{sc}
\end{scriptsize}
\end{center}
\end{table}

\begin{figure}[t] 
    \begin{center}
        \centerline{\includegraphics[width=1.0\linewidth]{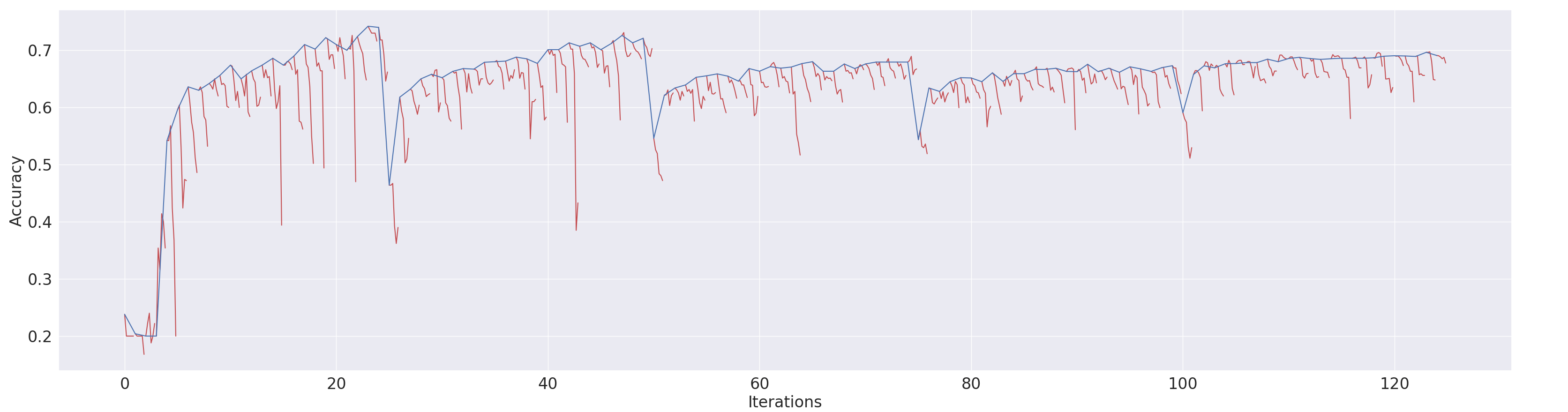}}
        \caption{Retained Accuracy (RA) for La-MAML plotted every 25 meta-updates up to Task 5 on CIFAR-100. RA at iteration \textit{j} (with \textit{j} increasing along the x-axis) denotes accuracy on all tasks seen uptil then. Red denotes the RA computed during the \emph{inner updates} (at $\theta_{k}^{j}$). Blue denotes RA computed at $\theta_{0}^{j+1}$ right after a \emph{meta-update}. We see that in the beginning, inner updates lead to catastrophic forgetting (CF) since the weights are not suitable for CL yet, but eventually become resistant when trained to retain old knowledge while learning on a stream of correlated data. We also see that RA maintains its value even as more tasks are added indicating that the model is successful at learning new tasks without sacrificing performance on old ones.}
        \label{fig:lookahead-plot}
    \end{center}
    \vskip -0.2in
\end{figure}

Table \ref{imgnet-table} reports the results of these experiments. We consistently observe superior performance of La-MAML as compared to other CL baselines on both datasets across setups. While the iCARL baseline attains lower BTI in some setups, it achieves that at the cost of much lower performance throughout learning. Among the high-performing approaches, La-MAML has the lowest BTI.
Recent work \cite{2019tinymemories, riemer2018learning} noted that Experience Replay (ER) is often a very strong baseline that closely matches the performance of the proposed algorithms. We highlight the fact that meta-learning and LR modulation combined show an improvement of more than 10 and 18\% (as the number of tasks increase from CIFAR to TinyImagenet) over the ER baseline in our case, with limited replay. 
Overall, we see that our method is robust and better-performing under both the standard and LLL setups of CL which come with different kinds of challenges. Many CL methods \cite{ebrahimi2019uncertainty, serra2018hat} are suitable for only one of the two setups. Further, as explained in Figure \ref{fig:lookahead-plot}, our model evolves to become resistant to \emph{forgetting} as training progresses. This means that beyond a point, it can keep making gradient updates on a small window of incoming samples without needing to do \emph{meta-updates}. 

\subsection{Evaluation of La-MAML's learning rate modulation}
\label{expt-modulation}
To capture the gains from learning the LRs, we compare La-MAML with our base algorithm, \textbf{C-MAML}. 
We ablate our choice of updating LRs asynchronously by constructing a version of C-MAML where per-parameter learnable LRs are used in the inner updates while the meta-update still uses a constant scalar LR during training. We refer to it as \emph{Sync-La-MAML} or \textbf{Sync} since it has synchronously updated LRs that don't modulate the meta-update. 
We also construct an ablation referred to as \emph{La-ER}, where the parameter updates are carried out as in ER but the LRs are modulated using the La-MAML objective's first-order version.
This tells us what the gains of LR modulation are over ER, since there is no meta-learning to encourage gradient alignment of the model parameters.
While only minor gains are seen on the MNIST benchmarks from asynchronous LR modulation, the performance gap increases as the tasks get harder. On CIFAR-100 and TinyImagenet, we see a trend in the RA of our variants with La-MAML performing best followed by \emph{Sync}. This shows that optimising the LRs aids learning and our \emph{asynchronous} update helps in knowledge consolidation by enforcing conservative updates to mitigate interference.

To test our LR modulation against an alternative \emph{bayesian} modulation scheme proposed in BGD \cite{bgd}, we define a baseline called Meta-BGD where per-parameter variances are modulated instead of LRs. This is described in further detail in Appendix \ref{baselines}. Meta-BGD emerges as a strong baseline and matches the performance of C-MAML given enough Monte Carlo iterations $m$, implying $m$ times more computation than C-MAML. Additionally, Meta-BGD was found to be sensitive to hyperparameters and required extensive tuning. We present a discussion of the robustness of our approach in Appendix \ref{robustness}, as well as a discussion of the setups adopted in prior work, in Appendix \ref{qualitative}.

We also compare the gradient alignment of our three variants along with ER in Table \ref{table-alignment} by calculating the cosine similarity between the gradients of the replay samples and newly arriving data samples. As previously stated, the aim of many CL algorithms is to achieve high gradient alignment across tasks to allow parameter-sharing between them. We see that our variants achieve an order of magnitude higher cosine similarity compared to ER, verifying that our objective promotes gradient alignment. 

\begin{table}[t] \sisetup{round-mode=places, round-precision=2 ,scientific-notation=false}
\caption{Gradient Alignment on CIFAR-100 and TinyImagenet dataset (values lie in [-1,1], higher is better)}
\label{table-alignment}
\vskip -0.1in
\begin{center}
\begin{small}
\begin{sc}
\begin{tabular}{l c c c c}
\toprule
Dataset  &              ER                   &                C-MAML                 &           SYNC                    &          La-MAML         \\
\midrule
CIFAR-100& \num{0.218e-2} {\tiny $\pm$ 0.0017} & \num{1.84e-2} {\tiny $\pm$ 0.0003} & \num{2.28e-2} {\tiny $\pm$ 0.0004} & \num{1.86e-2} {\tiny $\pm$ 0.0027} \\
TinyImagenet & \num{0.274e-2} {\tiny $\pm$ 0.0005} & \num{1.74e-2} {\tiny $\pm$ 0.0005} & \num{2.17e-2} {\tiny $\pm$ 0.0020} & \num{2.14e-2} {\tiny $\pm$ 0.0023} \\
\bottomrule
\end{tabular}
\end{sc}
\end{small}
\end{center}
\vskip -0.1in
\end{table}

\section{Conclusion}
We introduced La-MAML, an efficient meta-learning algorithm that leverages replay to avoid forgetting and favor positive backward transfer by learning the weights and LRs in an asynchronous manner. It is capable of learning online on a non-stationary stream of data and scales to vision tasks. We presented results that showed better performance against the state-of-the-art in the setup of efficient lifelong learning (LLL) \cite{chaudhry2018agem}, as well as the standard continual learning setting. 
In the future, more work on analysing and producing good optimizers for CL is needed, since many of our standard go-to optimizers like Adam \cite{2014adam} are primarily aimed at ensuring faster convergence in \emph{stationary} supervised learning setups.
Another interesting direction is to explore how the connections to \emph{meta-descent} can lead to more stable training procedures for meta-learning that can automatically adjust hyper-parameters on-the-fly based on training dynamics. 


\section*{Broader Impact}
This work takes a step towards enabling deployed models to operate while learning \emph{online}.
This would be very relevant for online, interactive services like recommender systems or home robotics, among others.
By tackling the problem of catastrophic forgetting, the proposed approach goes some way in allowing models to add knowledge incrementally without needing to be re-trained from scratch. Training from scratch is a compute intensive process, and even requires access to data that might not be available anymore. This might entail having to navigate a privacy-performance trade-off since many techniques like federated learning actually rely on not having to share data across servers, in order to protect user-privacy.

The proposed algorithm stores and replays random samples of prior data, and even with the higher alignment of the samples within a task under the proposed approach, there will eventually be some concept drift. While the proposed algorithm itself does not rely on or introduce any biases, any bias in the sampling strategy itself might influence the distribution of data that the algorithm \emph{remembers} and performs well on.

\begin{ack}

The authors are grateful to Matt Riemer, Sharath Chandra Raparthy, Alexander Zimin, Heethesh Vhavle and the anonymous reviewers for proof-reading the paper and suggesting improvements. 
This research was enabled in part by support provided by Compute Canada (\url{www.computecanada.ca}).

\end{ack}

{\small
\bibliography{neurips_2020}
}

\appendix

\newpage



\section{Hypergradient Derivation for La-MAML}
\label{hypergrad}
We derive the gradient of the weights $\theta^j_0$ and LRs $\alpha^j$ at time-step $j$ under the $k$-step MAML objective, with $L_t=\sum_{i=0}^{t} \ell_i$ as the \emph{meta-loss} and $\ell_t$ as the \emph{inner-objective}:
$
\begin{aligned}
\label{lr-gradient}
g_{\mathrm{MAML}}(\alpha^j) &=\frac{\partial}{\partial \alpha^j} L_{t}\left(\theta^j_{k}\right) =\frac{\partial}{\partial \theta^j_{k}} L_{t}\left(\theta^j_{k}\right) \cdot \frac{\partial}{\partial \alpha^j} \left(\theta^j_{k}\right) \nonumber \\  
&=\frac{\partial}{\partial \theta^j_{k}} L_{t}\left(\theta^j_{k}\right) \cdot \frac{\partial}{\partial \alpha^j} \left(U  \left(\theta^j_{k-1}\right)\right) \nonumber \\ 
&=\frac{\partial}{\partial \theta^j_{k}} L_{t}\left(\theta^j_{k}\right) \cdot \frac{\partial}{\partial \alpha^j} \left(\theta^j_{k-1}  - \alpha^j \frac{\partial \ell_{t} (\theta^j_{k-1})}{\partial \theta^j_{k-1}} \right) \nonumber \\  
&=\frac{\partial}{\partial \theta^j_{k}} L_{t}\left(\theta^j_{k}\right) \cdot \left( \frac{\partial}{\partial \alpha^j} \theta^j_{k-1}  - \frac{\partial}{\partial \alpha^j} \left( \alpha^j \frac{\partial \ell_{t} (\theta^j_{k-1})}{\partial \theta^j_{k-1}} \right)\right) \nonumber\\ 
&=\frac{\partial}{\partial \theta^j_{k}} L_{t}\left(\theta^j_{k}\right) \cdot \left( \frac{\partial}{\partial \alpha^j} \theta^j_{k-1}  -   \frac{\partial \ell_{t} (\theta^j_{k-1})}{\partial \theta^j_{k-1}} \right) \nonumber \\ 
& \quad \text {(Taking $ \frac{\partial \ell_{t} \left(\theta^j_{k-1}\right)}{\partial \theta^j_{k-1}}$ as a constant w.r.t $\alpha^j$ to get the first-order MAML approximation)  } \nonumber \\ 
&= \frac{\partial}{\partial \theta^j_{k}} L_{t} \left(\theta^j_{k}\right) \cdot \left( \frac{\partial}{\partial \alpha^j} U \left(\theta^j_{k-2}\right)  -  \left( \frac{\partial \ell_{t} (\theta^j_{k-1})}{\partial \theta^j_{k-1}} \right)\right) \nonumber \\ 
&=\frac{\partial}{\partial \theta^j_{k}} L_{t}\left(\theta^j_{k}\right) \cdot \left( \frac{\partial}{\partial \alpha^j} \theta^j_{0}  -  \sum_{k^{\prime}=0}^{k-1} \frac{\partial \ell_{t} (\theta^j_{n})}{\partial \theta^j_{n}}  \right) \quad (a) \nonumber \\ 
&=\frac{\partial}{\partial \theta^j_{k}} L_{t}\left(\theta^j_{k}\right) \cdot \left( -  \sum_{k^{\prime}=0}^{k-1} \frac{\partial \ell_{t} (\theta^j_{k^{\prime}})}{\partial \theta^j_{k^{\prime}}}  \right) \quad (b)
\end{aligned}
$

Where (a) is obtained by recursively expanding and differentiating the update function $U()$ as done in the step before it. (b) is obtained by assuming that the initial weight in the meta-update at time j : $\theta^j_{0}$, is constant with respect to $\alpha^j$.



Similarly we can derive the MAML gradient for the weights $\theta^j_0$, denoted as $g_{\mathrm{MAML}}(\theta^j_0)$ as: 

$
\begin{aligned}
g_{\mathrm{MAML}}(\theta^j_0) &=\frac{\partial}{\partial \theta^j_{0}} L_{t}(\theta^j_{k}) =\frac{\partial}{\partial \theta^j_{k}} L_{t}(\theta^j_{k}) \frac{\partial \theta^j_{k}}{\partial \theta^j_{0}} =\frac{\partial}{\partial \theta^j_{k}} L_{t}(\theta^j_{k})
\frac{\partial U_{k}(\theta^j_{k-1}) }{\partial \theta^j_{0}} \\ 
&=\frac{\partial}{\partial \theta^j_{k}} L_{t}(\theta^j_{k}) \frac{\partial}{\partial \theta^j_{k-1}} U(\theta^j_{k-1}) \cdots \frac{\partial}{\partial \theta^j_{0}} U(\theta^j_{1}) \\
& \quad \text { (repeatedly applying chain rule and using $\theta^j_k = U(\theta^j_{k-1})$ ) } \\ 
& =L_{t}^{\prime}(\theta^j_{k})
\left(I-\alpha \ell_{t}^{\prime \prime}(\theta^j_{k-1})\right) \cdots \left(I-\alpha \ell_{t}^{\prime \prime}(\theta^j_{0})\right) \\ 
& \quad\left(\text { using } U^{\prime}(\theta^j_{k'})=I-\alpha \ell_{t}^{\prime \prime}(\theta^j_{k'})\right) \quad \text { ($\textprime$  implies derivative with respect to argument) } \\ 
& = \left(\prod_{k^{\prime}=0}^{k-1}\left(I-\alpha \ell_{t}^{\prime \prime}(\theta^j_{k^{\prime}})\right)\right) L_{t}^{\prime}(\theta^j_{k}) \\ 
\end{aligned}\\
$
Setting all first-order gradient terms as constants to ignore second-order derivatives, we get the first order approximation as:

 $g_{\mathrm{FOMAML}}(\theta^j_0) = \left(\prod_{k^{\prime}=0}^{k-1}\left(I-\alpha \ell_{t}^{\prime \prime}\left(\theta^j_{k^{\prime}}\right)\right)\right) L_{t}^{\prime}(\theta^j_{k}) 
= L_{t}^{\prime}(\theta^j_{k}) $

In Appendix \ref{objective}, we show the equivalence of the C-MAML and CL objectives in Eq. \ref{our-objective} by showing that the gradient of the former ($g_{\mathrm{MAML}}(\theta^j_0)$) is equivalent to the gradient of the latter.

\newpage

\section{Equivalence of Objectives}
\label{objective}
It is straightforward to show that when we optimise the OML objective through the $k$-step MAML update, as proposed in C-MAML in Eq. \ref{eq:1}: 
\begin{equation}
    \min _{\theta^j_0} \mathbb{E}_{\tau_{1:t}}\left[L_{t}\left(U_{k}(\theta^j_0)\right)\right]
    \label{app-our-objective}
\end{equation}
where the \emph{inner-updates} are taken using data from the streaming task $\tau_t$, and the \emph{meta-loss} $L_t(\theta)=\sum_{i=1}^{t} \ell_i(\theta)$ is computed on the data from all tasks seen so far, it will correspond to minimising the following surrogate loss used in CL :

\begin{equation}
  \min _{\theta^j_0} \sum_{i=1}^{t} \left( \ell_{i}(\theta^j_0) - \alpha \frac{\partial \ell_{i}\left(\theta^j_0\right)}{\partial \theta^j_0} \cdot \frac{\partial \ell_{t}\left(\theta^j_0\right)}{\partial \theta^j_0} \right)
    \label{app-our-objective-surr}
\end{equation}

We show the equivalence for the case when $k=1$, for higher $k$ the form gets more complicated but essentially has a similar set of terms. Reptile
\cite{nichol2018first} showed that the $k$-step MAML gradient for the weights $\theta^j_0$ at time $j$, denoted as $g_{\mathrm{MAML}}(\theta^j_0)$ is of the form:

\begin{math}
\begin{aligned}
    {\frac{\partial L_{meta}(\theta^j_k)}{\partial \theta^j_0}} 
    & =  \bar{g}_{k}-\alpha \bar{H}_{k} \sum_{k^{\prime}=0}^{k-1} \bar{g}_{k^{\prime}}-\alpha \sum_{k^{\prime}=0}^{k-1} \bar{H}_{k^{\prime}} \bar{g}_{k}+O\left(\alpha^{2}\right) \quad \text{($\alpha$ is the \emph{inner-loop} learning rate)}\\
    & =  \bar{g}_{1}-\alpha \bar{H}_{1} \bar{g}_{0}-\alpha \bar{H}_{0} \bar{g}_{1}+O\left(\alpha^{2}\right) \quad \text{(using $k=1$)}\\
    & \text{Expressing the terms as derivatives, and using $ \frac{\partial}{\partial \theta^j_{0}}\left(\bar{g}_{0} \cdot \bar{g}_{1}\right) = \bar{H}_{1} \bar{g}_{0}+\bar{H}_{0} \bar{g}_{1} $, we get :} \\
    & = \frac{\partial  L_{meta}(\theta^j_{0})}{\partial \theta^j_{0}} - \frac{\partial}{\partial \theta^j_{0}}\left(\bar{g}_{0} \cdot \bar{g}_{1}\right) \\
    & = \frac{\partial \left(\sum_{i=1}^{t} \ell_{i}(\theta^j_0) -\alpha \bar{g}_{1} \cdot \bar{g}_{0} \right )}{\partial \theta^j_0} \quad \text{(substituting $L_{meta} = L_t = \sum_{i=1}^{t} \ell_i$)} \\
    & = \frac{\partial \left(\sum_{i=1}^{t} \ell_{i}(\theta^j_0) -\alpha \frac{\partial L_{meta}(\theta^j_0)}{\partial \theta^j_0} \frac{\partial \ell_{t}(\theta^j_0)}{\partial \theta^j_0}\right) }{\partial \theta^j_0}\\
    & = \frac{\partial \left(\sum_{i=1}^{t} \ell_{i}(\theta^j_0) -\alpha \frac{\partial\sum_{i=1}^{t}  \ell_{i}(\theta^j_0)}{\partial \theta^j_0}  \frac{\partial \ell_{t}(\theta^j_0)}{\partial \theta^j_0} \right )}{\partial \theta^j_0} \quad \text{(expanding $L_{meta}$)}\\
    & = \frac{\partial \left(\sum_{i=1}^{t} \ell_{i}(\theta^j_0) -\alpha \sum_{i=1}^{t} \frac{ \partial \ell_{i}(\theta^j_0)}{\partial \theta^j_0}  \frac{\partial \ell_{t}(\theta^j_0)}{\partial \theta^j_0} \right )}{\partial \theta^j_0} \\ 
    & \text{which is the same as the gradient of Eq. \ref{app-our-objective-surr}}.
\end{aligned}
\end{math}


where:
\begin{center}
\begin{alignat*}{3}
    \bar{g}_{k} 
    &= \frac{\partial L_{meta}\left(\theta^j_{0}\right)}{\partial \theta^j_{0}}  
    &&
    & &(\text{gradient of the \emph{meta-loss} evaluated at the initial point }) \\
    \bar{g}_{k^{\prime}} 
    &=\frac{\partial}{\partial \theta^j_{0}} L_{inner}(\theta^j_{0}) \quad &&(\text{for }k^{\prime}<k) \quad
    & &(\text{gradients of the \emph{inner-updates} evaluated at the initial point}) \\
    \theta^j_{k^{\prime}+1} 
    &=\theta^j_{k^{\prime}}-\alpha g_{k^{\prime}}
    &&
    & &(\text{sequence of parameter vectors})
\end{alignat*}
\end{center}

\begin{center}
\begin{alignat*}{3}
    \bar{H}_{k} 
    &=L_{meta}^{\prime \prime}\left(\theta^j_{0}\right) 
    &&
    & &\text {(Hessian of the \emph{meta-loss} evaluated at the initial point) }  \\
    \bar{H}_{k^{\prime}} 
    &=L_{inner}^{\prime \prime}\left(\theta^j_{0}\right)  \quad
    &&(\text{for } k^{\prime}<k)  \quad
    & &\text{(Hessian of the \emph{inner-objective} evaluated at the initial point)}  \\
    \text{And,} &\text{ in our case:} \\
    L_{meta} 
    &= L_t=\sum_{i=1}^{t} \ell_i \\ 
    L_{inner} 
    &= \ell_t
\end{alignat*}
\end{center}

\textbf{Bias in the objective}: We can see in Eq. \ref{app-our-objective-surr} that the gradient alignment term introduces some bias, which means that the parameters don't exactly converge to the minimiser of the losses on all tasks. This has been acceptable in the CL regime since we don't aim to reach the minimiser of some stationary distribution anyway (as also mentioned in Section \ref{other-work}). If we did converge to the minimiser of say $t$ tasks at some time $j$, this minimiser would no longer be optimal as soon as we see the new task $\tau_{t+1}$.
    Therefore, in the limit of infinite tasks and time, ensuring low-interference between tasks will pay off much more as opposed to being able to converge to the exact minima, by allowing us to make shared progress on both previous and incoming tasks.

\section{C-MAML Algorithm}
\label{cmamlalgo}
Algorithm \ref{alg:cmaml} outlines the training procedure for the C-MAML algorithm  we propose \footnote{Our algorithm, \emph{Continual-MAML} is different from a concurrent work \url{https://arxiv.org/abs/2003.05856} which proposes an algorithm with the same name}.

\begin{algorithm}[H] 
   \caption{C-MAML}
   \label{alg:cmaml}
\begin{algorithmic}
    \State {\bfseries Input:} Network weights $\theta^0_0$, inner objective $\ell$, meta objective $L$, Inner learning rate $\alpha$, Outer learning rate $\beta$
    \State $j \leftarrow 0$
    \State $R \leftarrow \{ \}$ \Comment{Initialise replay-buffer}
    
    \For{$t:=1$ {\bfseries to} $T$}
    \State $(X^t,Y^t) \sim D_t$

    \For{$ep:=1$ {\bfseries to} $num_{epochs}$}
   
        \For{batch $b$ {\bfseries in} $(X^t,Y^t)$}
            \State $k \leftarrow sizeof(b)$          
           \State $b_m \leftarrow Sample(R) \cup b$ \Comment{batch of samples from $\tau_{1:t}$ for \emph{meta-loss}}
           
           \For{$k^{\prime}=0$ {\bfseries to} $k-1$}
                \State Push $b[k^{\prime}]$ to R with some probability based on reservoir sampling
                
                \State $\theta_{k^{\prime}+1}^{j} \leftarrow  \theta_{k^{\prime}}^{j}  - \alpha \cdot \nabla_{\theta_{k^{\prime}}^{j}} \ell_{t}(\theta_{k^{\prime}}^{j}, b[k^{\prime}])$ \Comment{\emph{inner-update} on each incoming sample}
           \EndFor
            \State $\theta^{j+1}_{0} \leftarrow \theta_0^{j} - \beta \cdot \nabla_{\theta_0^{j}}  L_{t}(\theta_k^j, b_m)$  \Comment{\emph{outer-update} by differentiating \emph{meta-loss}}
            \State   $j \leftarrow j+1$ 
        \EndFor
    \EndFor
   \EndFor
\end{algorithmic}
\end{algorithm}

\section{Inter-Task Alignment}
\label{representation}

We assume that at time $j$ during training, we are seeing samples from the streaming task $\tau_t$.
It is intuitive to realise that incentivising the alignment of all $\tau_{1:t}$ with the current $\tau_{t}$ indirectly also incentivises the alignment amongst $\tau_{1:t-1}$ as well.
To demonstrate this, we compute the mean dot product of the gradients amongst the old tasks $\tau_{1:t-1}$ as the new task $\tau_t$ is added, for $t$ varying from 2 to 11. We do this for C-MAML and La-MAML on CIFAR-100. 

\begin{figure}[H] 
        \centering
    \begin{subfigure}{.499\linewidth}
      \centering
      \includegraphics[width=1.0\linewidth]{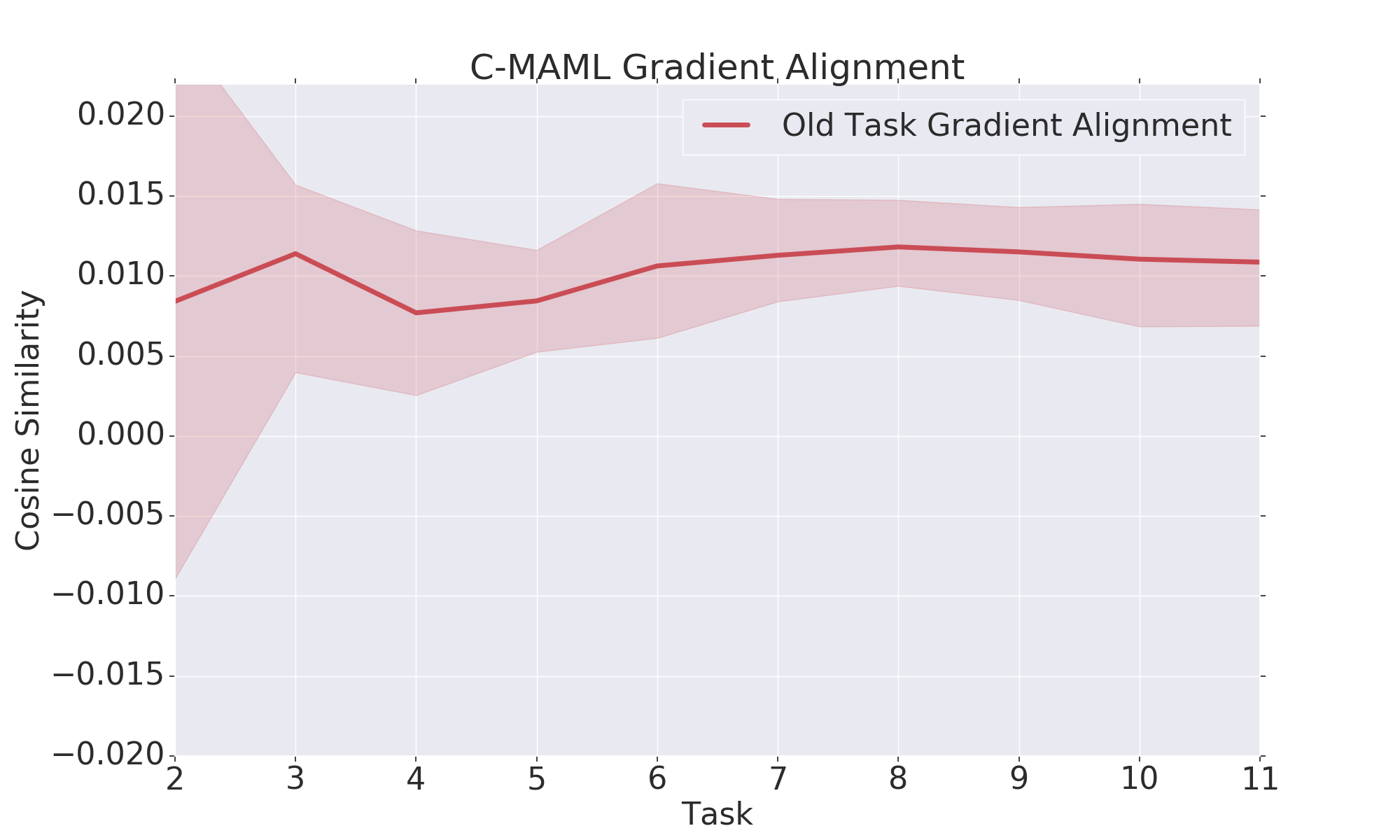}
      \caption{C-MAML}
      \label{align-old-tasks-cmaml}
    \end{subfigure}%
    \hfill
    \begin{subfigure}{.499\linewidth}
      \centering
      \includegraphics[width=1.0\linewidth]{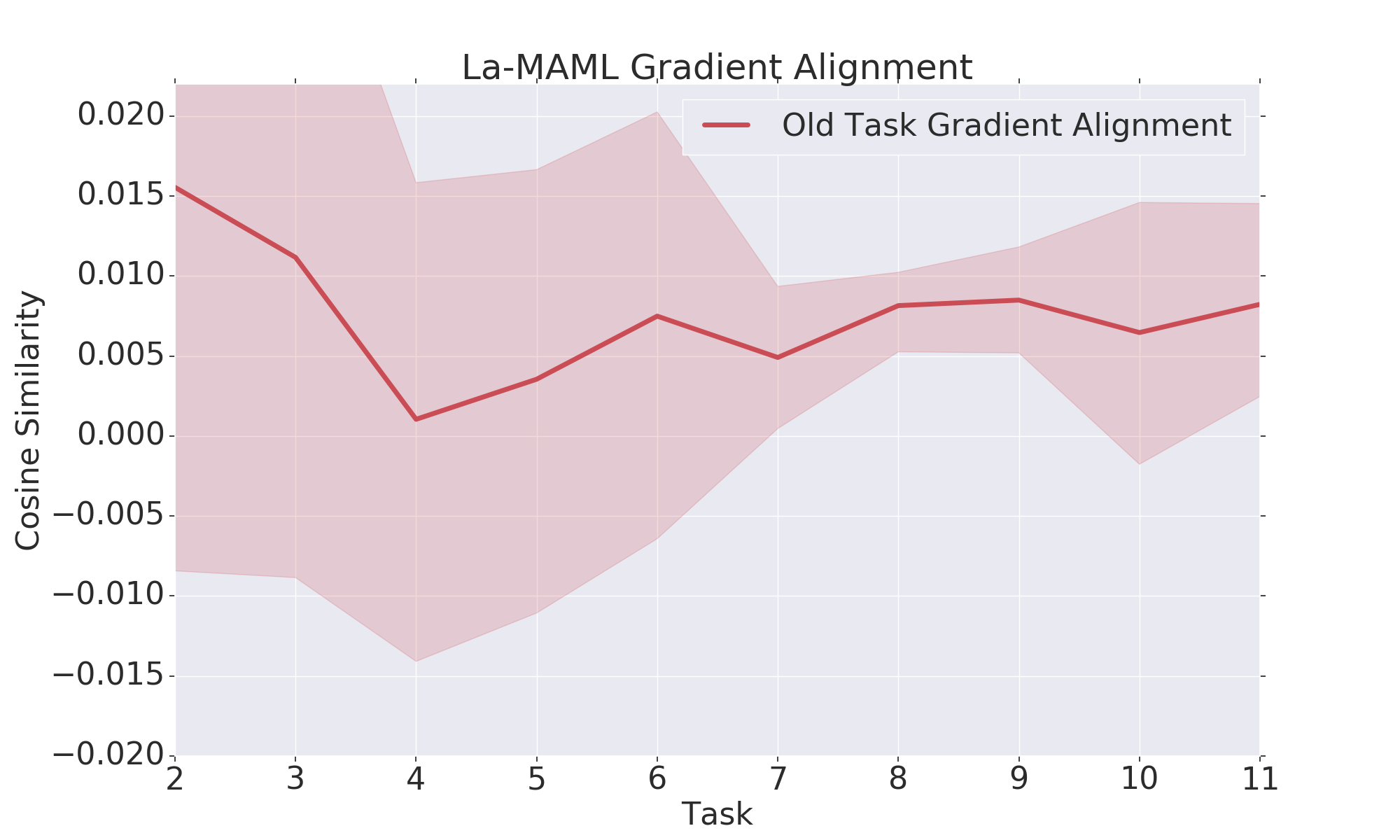}
      \caption{La-MAML}
      \label{align-old-tasks-lamaml}
    \end{subfigure}
    \caption{Average dot product amongst gradients of $\tau_{1:t-1}$ as new tasks are added, for the C-MAML and La-MAML algorithms calculated over 5 runs. \emph{x-axis} shows the streaming task ID, $t$ and \emph{y-axis} shows the cosine similarity.} 
\end{figure}

As can be seen in Figures \ref{align-old-tasks-cmaml} and \ref{align-old-tasks-lamaml}, the alignment stays positive and roughly constant even as more tasks are added. 


\section{Robustness}
\label{robustness}
Learning rate is one of the most crucial hyper-parameters during training and it often has to be tuned extensively for each experiment. In this section we analyse the robustness of our proposed variants to their LR-related hyper-parameters on the CIFAR-100 dataset. 
Our three variants have different sets of these hyper-parameters which are specified as follows: 
\begin{itemize}
    \item \textbf{C-MAML}: Inner and outer update LR (scalar) for the weights ($\alpha$ and $\beta$)
    \item \textbf{Sync La-MAML}: Inner loop initialization value for the vector LRs ($\alpha_0$), scalar learning rate of LRs ($\eta$) and scalar learning rate for the weights in the outer update ($\beta$)
    \item \textbf{La-MAML}: Scalar initialization value for the vector LRs ($\alpha_0$) and a scalar learning rate of LRs ($\eta$)
\end{itemize}

\begin{figure}[ht]
    \centering
    \begin{subfigure}{.5\linewidth}
      \centering
      \includegraphics[width=1.0\linewidth]{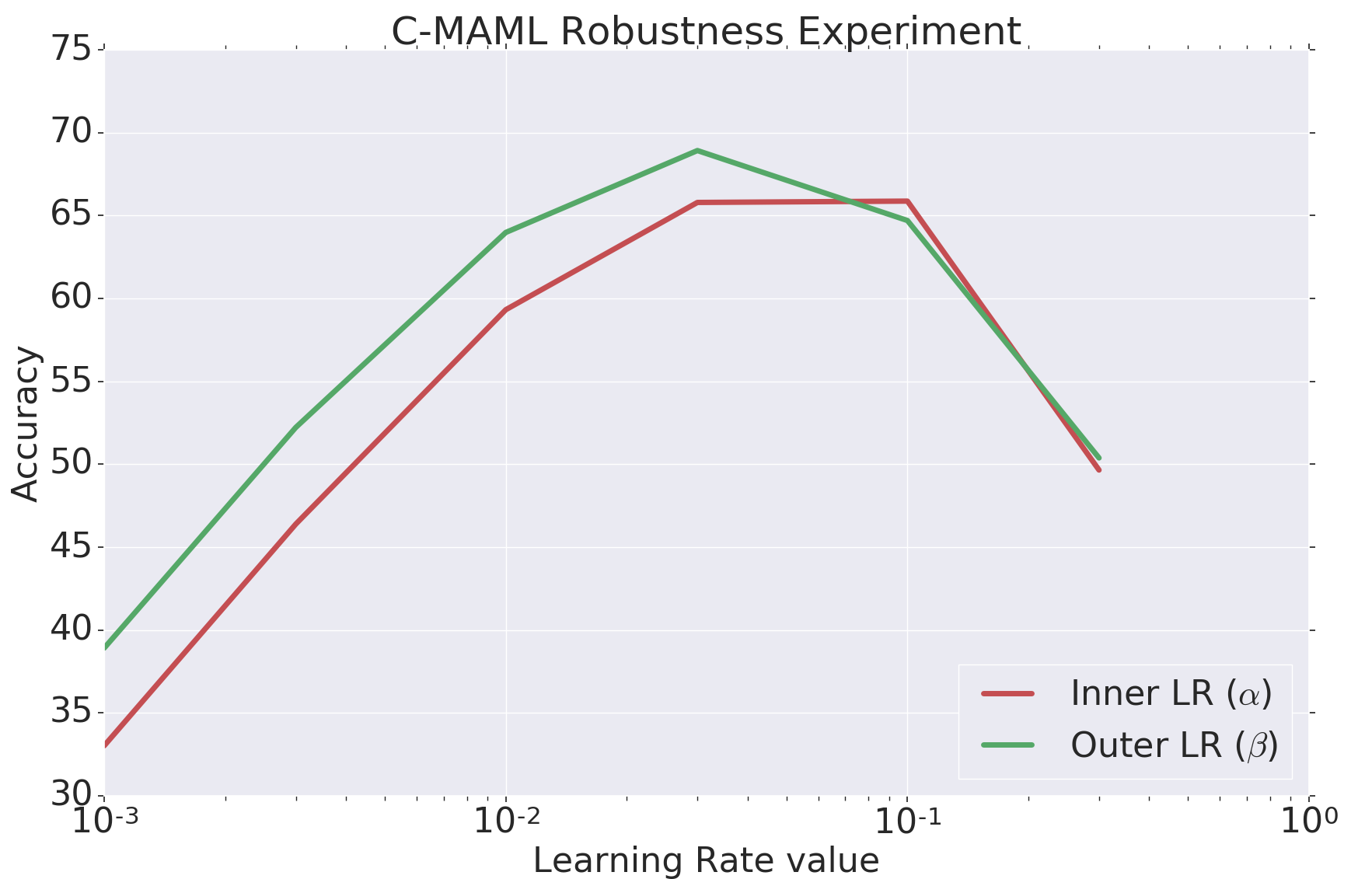}
      \caption{C-MAML: Modulation of $\alpha$ and $\beta$}
      \label{params:cmaml}
    \end{subfigure}%
    \begin{subfigure}{.5\linewidth}
      \centering
      \includegraphics[width=1.0\linewidth]{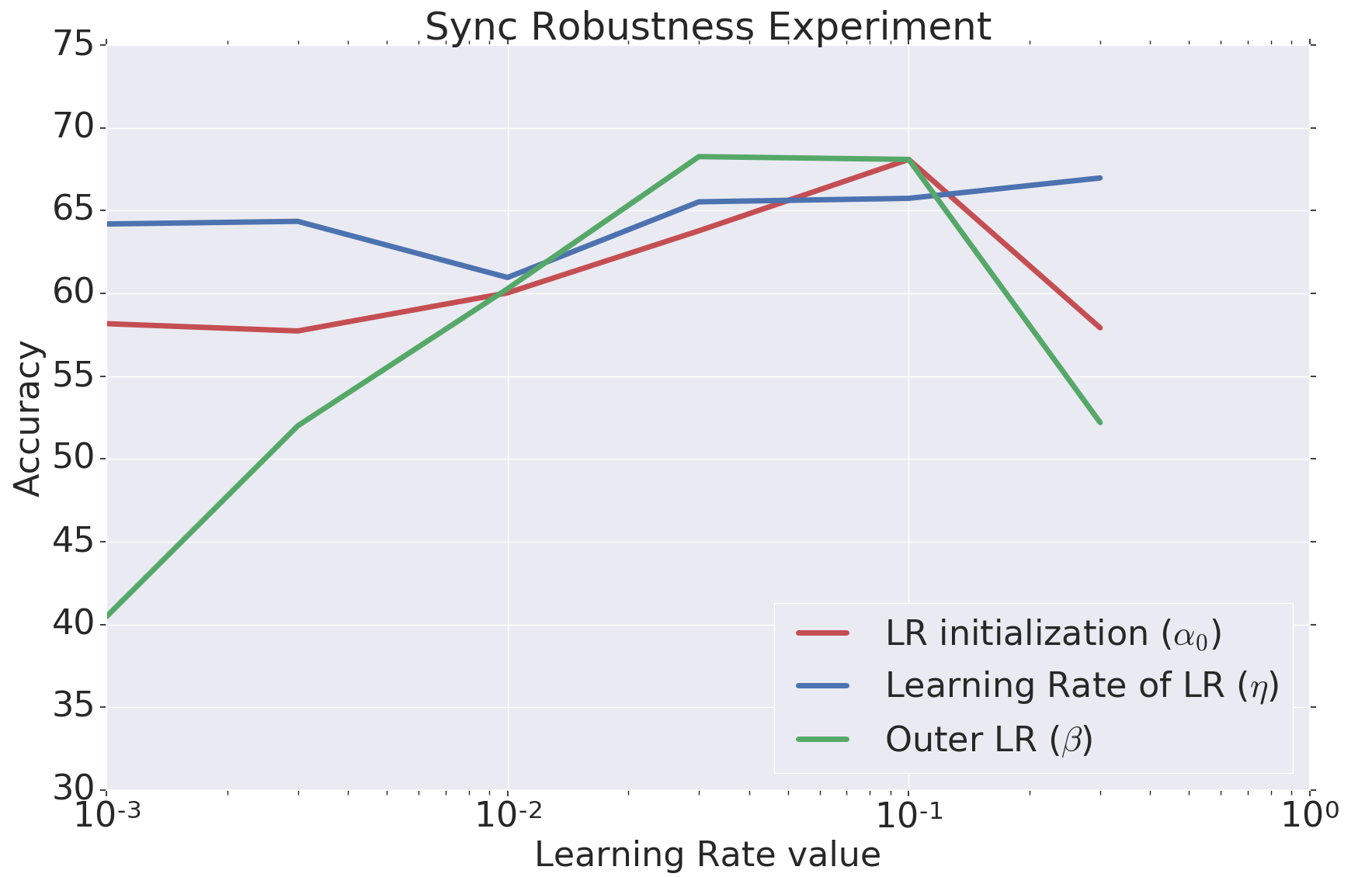}
      \caption{Sync: Modulation of $\alpha_0$, $\eta$ and $\beta$}
      \label{params:sync}
    \end{subfigure}
    \medskip
    \begin{subfigure}{.5\linewidth}
      \centering
      \includegraphics[width=1.0\linewidth]{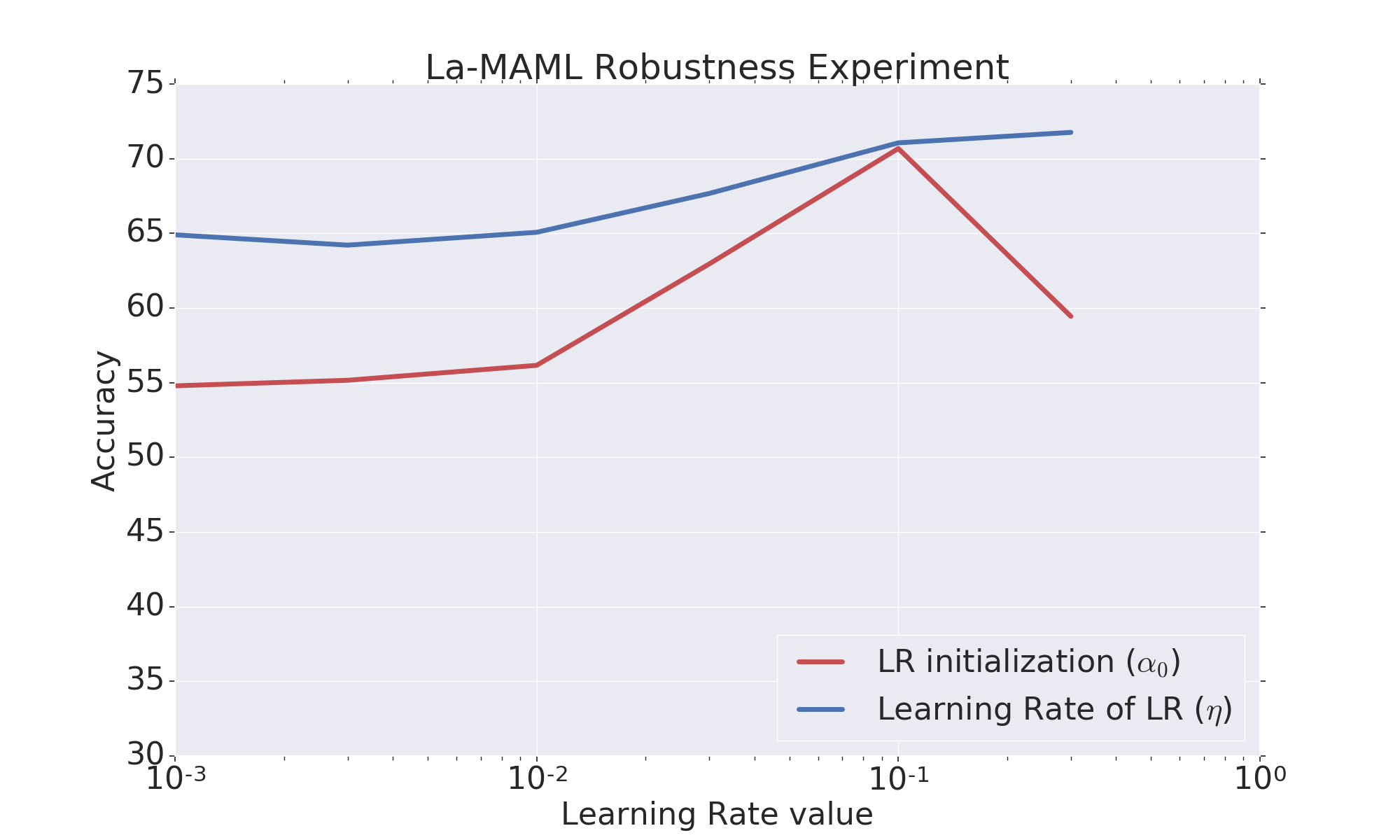}
      \caption{La-MAML: Modulation of $\alpha_0$ and $\eta$}
      \label{params:lamaml}
    \end{subfigure}
    \caption{Retained Accuracy vs Learning Rates plot for La-MAML and its variants. Figures are plotted by varying one of the learning rate hyperparameter while keeping the others fixed at their optimal value. The hyperparameter is varied between [0.001, 0.3].}
    \label{timing}
\end{figure}

La-MAML is considerably more robust to tuning compared to its variants, as can be seen in Figure \ref{params:lamaml}. We empirically observe that it only requires tuning of the initial value of the LR, while being relatively insensitive to the learning rate of the LR ($\eta$). We see a consistent trend where the increase in $\eta$ leads to an increase in the final accuracy of the model. This increase is very gradual, since across a wide range of LRs varying over 2 orders of magnitude (from 0.003 to 0.3), the difference in RA is only 6\%. This means that even without tuning this parameter ($\eta$), La-MAML would have outperformed most baselines at their optimally tuned values.

As seen in Figure \ref{params:cmaml}, C-MAML sees considerable performance variation with the tweaking of both the inner and outer LR. We also see that the effects of the variations of the inner and outer LR follow very similar trends and their optimal values finally selected are also identical. This means that we could potentially tune them by doing just a 1D search over them together instead of varying both independently through a 2D grid search.
The Sync version of La-MAML (Figure \ref{params:sync}), while being relatively insensitive to the scalar initial value $\alpha_0$ and the $\eta$, sees considerable performance variation as the outer learning rate for the weights: $\beta$ is varied. This variant has the most hyper-parameters and only exists for the purpose of ablation.

\begin{figure}[ht]
    \centering
    \begin{subfigure}{.499\linewidth}
      \centering
      \includegraphics[width=1.0\linewidth]{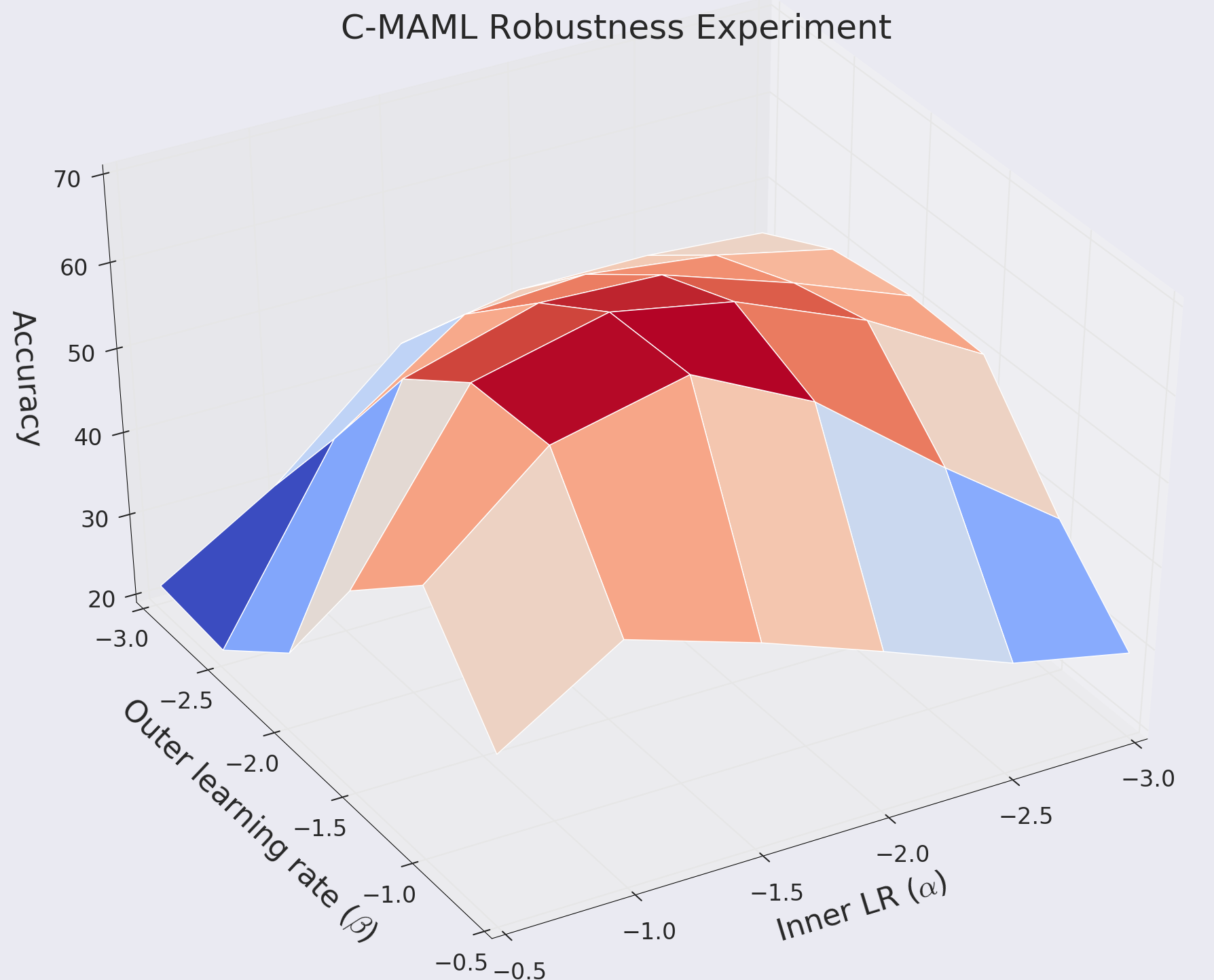}
      \caption{C-MAML: Modulation of $\alpha$ and $\beta$}
      \label{fig:sub1}
    \end{subfigure}%
    \hfill
    \begin{subfigure}{.499\linewidth}
      \centering
      \includegraphics[width=1.0\linewidth]{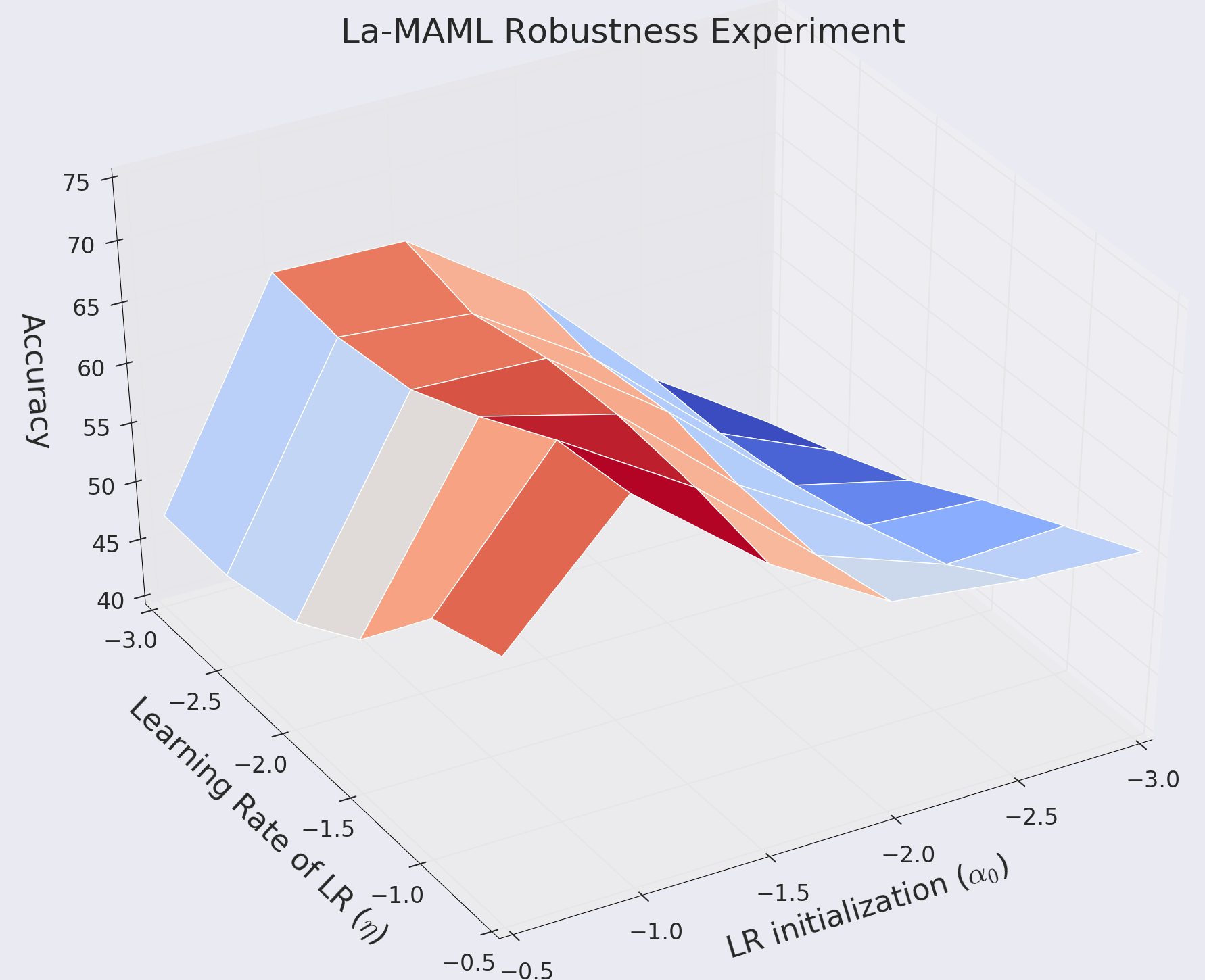}
      \caption{La-MAML: Modulation of $\alpha_0$ and $\eta$}
      \label{fig:sub2}
    \end{subfigure}
    \caption{Plots of Retained Accuracy (RA) across hyper-parameter variation for C-MAML and La-MAML. We show results of the grid search over the learning rate hyperparameters. RA decreases from red to blue. All the hyperparameters are varied between [0.001, 0.3], with the axes being in log-scale.}
    \label{3d-robustness}
\end{figure}

Fig. \ref{3d-robustness} shows the result of 2D grid-searches over sets of the above-mentioned hyper-parameters for C-MAML and La-MAML for a better overview.

\section{Timing Comparisons}
In this section, we compare the wall-clock running times (\emph{Retained Accuracy} (RA) versus \emph{Time}) of La-MAML against other baselines on the CIFAR100 dataset in the multi-pass setting. For ER, iCarl and La-MAML we see an increasing tread in the RA vs Time plot with La-MAML having the best RA at the expense of the increase in time. In contrast, both AGEM and GEM  perform worse than La-MAML while also taking much more running time.



\begin{figure}[ht]
\vskip 0.2in
\begin{center}
\centerline{\includegraphics[width=0.5\linewidth]{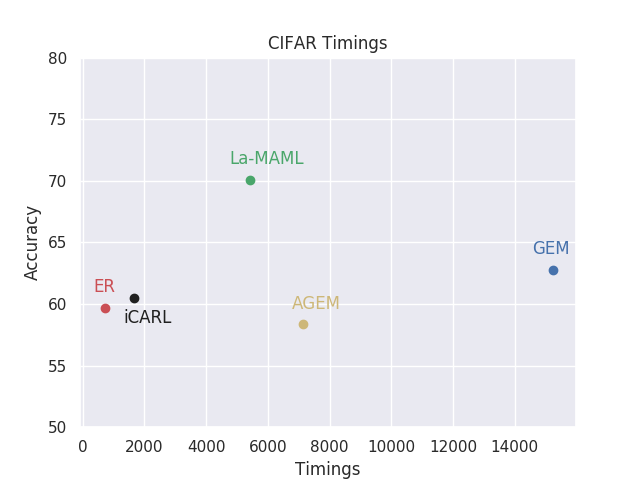}}
\caption{Retained Accuracy vs Running time (seconds) for La-MAML vs other baselines on the CIFAR100 dataset.}
\label{timing}
\end{center}
\vskip -0.2in
\end{figure}

\section{Experimental}
\label{experimental}

We carry out hyperparameter tuning for all the approaches by performing a grid-search over the range [0.0001 - 0.3] for hyper-parameters related to the learning-rate. For the multi-pass setup we use 10 epochs for all the CL approaches. In the single pass setup, all compared approaches have a hyper-parameter called \emph{glances} which indicates the number of gradient updates or meta-updates made on each incoming sample of data. In the \emph{Single-Pass} (LLL) setup, it becomes essential to take multiple gradient steps on each sample (or see each sample for multiple \emph{glances}), since once we move on to later samples, we can't revisit old data samples. The performance of the algorithms naturally increases with the increase in \emph{glances} up to a certain point. To find the optimal number of \emph{glances} to take over each sample, we search over the values [1,2,3,5,10]. Tables \ref{params-table} and \ref{params-table-mnist} lists the optimal hyperparameters for all the compared approaches. All setups used the SGD optimiser since it was found to preform better than Adam \cite{2014adam} (possibly due to reasons stated in Section \ref{other-work} regarding the CL setup). 

To avoid exploding gradients, we clip the gradient values of all approaches at a norm of 2.0. \emph{Class divisions} across different tasks vary with the random seeds with which the experiments were conducted. Overall, we did not see much variability across different class splits, with the variation being within 0.5-2\% of the mean reported result as can be seen from Table \ref{imgnet-table}

For all our baselines, we use a constant batch-size of 10 samples from the streaming task. This batch is augmented with 10 samples from the replay buffer for the replay-based approaches. 
La-MAML and its variants split the batch from the streaming task into a sequence of smaller disjoint sets to take multiple ($k=10$ for MNIST and $k=5$ for CIFAR100/TinyImagenet) gradient steps in the \emph{inner-loop}. In MER, each sample from the incoming task is augmented with a batch of 10 replay samples to form the batch used for the meta-update. We found very small performance gaps between the first and second-order versions of our proposed variants with performance differences in the range of 1-2\% for RA. This is in line with the observation that deep neural networks have near-zero hessians since the ReLU non-linearity is linear almost everywhere \cite{zero-hessian}.


\newcommand\Tstrut{\rule{0pt}{2.6ex}} 
\begin{table}[t]
    \caption{Final hyperparameters for all compared approaches on the CIFAR and TinyImagenet benchmarks}
    \label{params-table}
    \begin{center}
    \begin{small}
    \begin{sc}
    \begin{tabular}{lccccc}
        \toprule
        Method      &       Parameter        & \multicolumn{2}{c}{CIFAR-100}  & \multicolumn{2}{c}{TinyImagenet}     \\
        \cmidrule{3-4}  \cmidrule{5-6} 
                    &                        &   Single     &    Multiple     &     Single     &    Multiple     \\
        \midrule
        ER          & \emph{LR}              &     0.03     &       0.03      &         0.1    &      0.1        \\
                    & \emph{Epochs/Glances}  &       10     &       10        &         10     &       10        \\  \hline \Tstrut
        IID         & \emph{LR}              &       -      &       0.03      &         -      &       0.01      \\ 
                    & \emph{Epochs/Glances}  &      -       &       50        &         -      &       50        \\ \hline \Tstrut
        iCARL       & \emph{LR}              &     0.03     &         0.03    &       0.01     &     0.01        \\
                    & \emph{Epochs/Glances}  &       2      &       10        &         2      &       10        \\ \hline \Tstrut
        GEM         & \emph{LR}              &     0.03     &        0.03     &      0.03      &       0.03      \\
                    & \emph{Epochs/Glances}  &       2      &       10        &         2      &       10        \\ \hline \Tstrut
        AGEM        & \emph{LR}              &     0.03     &        0.03     &     0.01       &       0.01      \\
                    & \emph{Epochs/Glances}  &       2      &       10        &         2      &       10        \\ \hline \Tstrut
        MER         & \emph{LR $\alpha$}     &     0.1      &        -        &      0.1       &       -         \\
                    & \emph{LR $\beta$}      &      0.1     &       -         &      0.1       &       -         \\ 
                    & \emph{LR $\gamma$}     &       1      &       -         &      1         &       -         \\
                    & \emph{Epochs/Glances}  &       10     &       -         &       10       &       -         \\ \hline \Tstrut
        Meta-BGD    & \emph{$\eta$}          &     50       &          50     &       50       &        -        \\
                    & \emph{std-init}        &     0.02     &        0.02     &       0.02     &        -        \\
                    & \emph{$\beta_{inner}$} &     0.1      &         0.1     &      0.1       &        -        \\
                    & \emph{mc-iters}        &       2      &        2        &         2      &        -        \\
                    & \emph{Epochs/Glances}  &       3      &       10        &         3      &        -        \\ \hline \Tstrut
        C-MAML     & \emph{$\alpha$}         &     0.03     &       0.03       &     0.03       &      0.03       \\
                    & \emph{$\beta$}         &     0.03     &       0.03      &    0.03        &      0.03       \\
                    & \emph{Epochs/Glances}  &       5      &       10        &         2      &       10        \\ \hline \Tstrut    
        La-ER       & \emph{$\alpha_0$}      &     0.1      &       0.1       &     0.03               &    0.03            \\
                    & \emph{$\eta$}          &     0.1      &       0.1       &     0.1    &    0.1          \\
                    & \emph{Epochs/Glances}  &       1      &       10        &         2      &       10        \\ \hline \Tstrut
        Sync La-MAML& \emph{$\alpha_0$}      &     0.1      &       0.1       &       0.075    &  0.075          \\
                    & \emph{$\beta$}         &     0.1      &       0.1       &       0.075    &    0.075        \\
                    & \emph{$\eta$}          &     0.3      &      0.3        &     0.25       &       0.25      \\
                    & \emph{Epochs/Glances}  &       5      &       10        &         2      &       10        \\ \hline \Tstrut
        La-MAML     & \emph{$\alpha_0$}      &     0.1      &        0.1      &   0.1          &  0.1            \\
                    & \emph{$\eta$}          &     0.3      &       0.3       &     0.3        &    0.3          \\
                    & \emph{Epochs/Glances}  &       10     &       10        &         2      &       10        \\


        \bottomrule
    \end{tabular}
    \end{sc}
    \end{small}
    \end{center}
    \vskip -0.1in
\end{table}

\begin{table}[t]
    \caption{Final hyperparameters used for our variants on the MNIST benchmarks}
    \label{params-table-mnist}
    \begin{center}
    \begin{small}
    \begin{sc}
    \begin{tabular}{lccccc}
        \toprule
        Method      &       Parameter        & Permutations  & Rotations & Many                     \\
        \midrule
         C-MAML     & \emph{$\alpha$}        &     0.03     &       0.1      &         0.03         \\
                    & \emph{$\beta$}         &     0.1      &       0.1      &         0.15         \\
                    & \emph{Glances}         &      5       &       5         &         5           \\ \hline \Tstrut    
                    
        Sync La-MAML& \emph{$\alpha_0$}      &     0.15      &       0.15       &         0.03       \\
                    & \emph{$\beta$}         &     0.1      &       0.3       &         0.03       \\
                    & \emph{$\eta$}          &     0.1      &      0.1        &         0.1        \\
                    & \emph{Glances}         &      5       &       5         &         10          \\ \hline \Tstrut
                    
        La-MAML     & \emph{$\alpha_0$}      &     0.3      &        0.3      &         0.1          \\
                    & \emph{$\eta$}          &     0.15      &       0.15       &         0.1         \\
                    & \emph{Glances}         &       5      &       5         &         10          \\ 
                    
        \bottomrule
    \end{tabular}
    \end{sc}
    \end{small}
    \end{center}
    \vskip -0.1in
\end{table}

\textbf{MNIST Benchmarks}:
On the MNIST continual learning benchmarks, images of size 28x28 are flattened to create a 1x784 array. This array is passed on to a fully-connected neural network having two layers with 100 nodes each. Each layer uses ReLU non-linearity. The output layer uses a single head with 10 nodes corresponding to the 10 classes. In all our experiments, we use a modest replay buffer of size 200 for MNIST Rotations and Permutation and size 500 for Many Permutations.  

\textbf{Real-world visual classification}:
For CIFAR and TinyImageNet we used a CNN having 3 and 4 conv layers respectively with 160 3x3 filters. The output from the final convolution layer is flattened and is passed through 2 fully connected layers having 320 and 640 units respectively. All the layers are succeeded by ReLU nonlinearity. Finally, a multi-headed output layer is used for performing 5-way classification for every task. This architecture is used in prior meta-learning work \cite{yu2020gradient}.

For CIFAR and TinyImagenet, we allow a replay buffer of size 200 and 400 respectively which implies that each class in these dataset gets roughly about 1-2 samples in the buffer. 
For TinyImagenet, we split the validation set into \emph{val} and \emph{test} splits, since the labels in the actual test set are not released.

\section{Baselines}
\label{baselines}
On the MNIST benchmarks, we compare our algorithm against the baselines used in \cite{riemer2018learning}, which are as follows:

\begin{itemize}
\itemsep0.5em
    \item Online: A baseline for the LLL setup, where a single network is trained one example at a time with SGD.
    \item EWC \cite{Kirkpatrick3521}: Elastic Weight Consolidation is a regularisation based method which constraints the weights important for the previous tasks to avoid catastrophic forgetting.
    \item GEM \cite{lopez2017gradient}: Gradient Episodic Memory does constrained optimisation by solving a quadratic program on the gradients of new and replay samples, trying to make sure that these gradients do not alter the past tasks' knowledge.
    \item MER \cite{riemer2018learning}: Meta Experience Replay samples i.i.d data from a replay memory to meta-learn model parameters that show increased gradient alignment between old and current samples. We evaluate against this baseline only in the LLL setups.
\end{itemize}

On the real-world visual classification dataset, we carry out experiments on GEM, MER along with:-
\begin{itemize}
\itemsep0.5em
    \item IID: Network gets the data from all tasks in an independent and identically distributed manner, thus bypassing the issue of catastrophic forgetting completely. Therefore, IID acts as an upper bound for the RA achievable with this network.
    \item ER: Experience Replay uses a small replay buffer to store old data using reservoir sampling. This stored data is then replayed again along with the new data samples.
    \item iCARL \cite{rebuffi2017icarl}: iCARL is originally from the family of class incremental learners, which learns to classify images in the metric space. It prevents catastrophic forgetting by using a memory of exemplar samples to perform distillation from the old network weights. Since we perform experiments in a task incremental setting, we use the modified version of iCARL (as used by GEM \cite{lopez2017gradient}), where distillation loss is calculated only over the logits of the particular task.
    \item A-GEM \cite{chaudhry2018agem}: Averaged Gradient Episodic Memory proposed to project gradients of the new task to a direction such as to avoid interference with respect to the average gradient of the old samples in the buffer. 
    \item Meta-BGD: Bayesian Gradient Descent \cite{bgd} proposes training a bayesian neural network for CL where the learning rate for the parameters (the means) are derived from their variances. We construct this baseline by equipping C-MAML with bayesian training, where each parameter in $\theta$ is now sampled from a gaussian distribution with a certain mean and variance.
    The inner-loop stays same as C-MAML(constant LR), but the magnitude of the meta-update to the parameters in $\theta$ is now influenced by their associated variances. The variance updates themselves have a closed form expression which depends on $m$ monte-carlo samples of the \emph{meta-loss}, thus implying $m$ forward passes of the inner-and-outer loops (each time with a newly sampled $\theta$) to get $m$ meta-gradients. 
\end{itemize}

\section{Discussion on Prior Work}
\label{qualitative}
In Table \ref{table:setups}, we provide a comparative overview of various continual learning methods to situate our work better in the context of prior work. 

\emph{Prior-focused methods} face model capacity saturation as the number of tasks increase. These methods freeze weights to defy forgetting, and so penalise changes to the weights, even if those changes could potentially improve model performance on old tasks.
They are also not suitable for the LLL setup (section \ref{experiments}), since it requires many passes through the data for every task to learn weights that are optimal enough to be frozen.
Additionally, the success of weight freezing schemes can be attributed to over-parameterisation in neural networks, leading to sub-networks with sufficient capacity to learn separate tasks. However continual-learning setups are often motivated in resource-constrained settings requiring efficiency and scalability. Therefore solutions that allow light-weight continual learners are desirable. Meta-learning algorithms are able to exploit even small models to learn a good initialization where gradients are aligned across tasks, enabling shared progress on optimisation of task-wise objectives. Our method additionally allows meta-learning to also achieve a prior-focusing affect through the async-meta-update, without necessarily needing over-parameterised models.

In terms of resources, \emph{meta-learning based methods} require smaller replay memories than traditional methods because they learn to generalise better across and within tasks, thus being sample-efficient. Our learnable learning rates incur a memory overhead equal to the parameters of the network. This is comparable to or less than many prior-based methods that store between 1 to $T$ scalars per parameter depending on the approach ($T$ is the number of tasks). 

It should be noted that our learning rate modulation involves clipping updates for parameters with non-aligning gradients. In this aspect, it is related to methods like GEM and AGEM mentioned before. Where the distinction lies, is that our method takes some of the burden off of the clipping, by ensuring that gradients are more aligned in the first place. This means that there should be less interference and therefore less clipping of updates deemed essential for learning new tasks, on the whole.

\begin{table*}[hb]
\centering
\caption{\textbf{Setups in prior work:} We describe the setups and assumptions adopted by prior work, focusing on approaches relevant to our method. FWT and BWT refer to forward and backward transfer as defined in \cite{lopez2017gradient}. '-' refers to no inductive bias for or against the specific property.
Saturation of capacity refers to reduced network plasticity due to weight change penalties gradually making further learning impossible. 
The LLL setup is defined in Section \ref{experiments}.
$<$ and $>$ with replay indicate that a method's replay requirements are lesser or more compared to other methods in the table. \emph{Fishers} refers to the Fisher Information Matrix (FIM) computed per task. Each FIM has storage equal to that of the model parameters. Approaches using Bayesian Neural Networks require twice as many parameters (as does La-MAML) to store the mean and variance estimates per parameter.  
}
\begin{center}
    \begin{scriptsize}
    \begin{sc}
  \begin{small}
    \resizebox{0.85\textwidth}{!}{%
        \begin{tabular}{c|cc|c|cc|c}
        \textbf{Approach}            &  \multicolumn{2}{c|}{\textbf{Transfer}}     & \multicolumn{1}{c|}{\textbf{Capacity}}             & \multicolumn{2}{c|}{\textbf{Resources}} & \textbf{Algorithm} \\
                    &  FWT & BWT    & Saturates &  LLL  & Storage  &      \\\hline
        Prior-Focused        & - & $\times$ & $\surd$            & $\times$ & T Fishers  &  EWC \cite{Kirkpatrick3521}   \\  
        Prior Focused       & - & $\times$ & $\surd$           & $\times$  & T masks     & HAT \cite{serra2018hat} \\
        Prior Focused       & - & $\times$ & $\surd$           & $\surd$  & 2x params     & BGD/UCB \cite{bgd} \cite{ebrahimi2019uncertainty}\\

        Replay              & -  &    -  & $\times$             & $\surd$  &  $>$ replay    & iCARL \cite{rebuffi2017icarl} \\ 
        Replay              & - &     - & $\times$             & $\surd$  & $>$ replay    & GEM \cite{lopez2017gradient} \\

        Meta + Replay           & $\surd$ & $\surd$             &  $\times$         & $\surd$  & replay    & MER \cite{riemer2018learning} \\
        Meta + Replay            & $\surd$ & $\surd$             &  $\times$    & $\surd$  & replay & Ours \\

        \end{tabular}
    }
    \end{small}
   \end{sc}
    \end{scriptsize}
\end{center}
\label{table:setups}
\end{table*}

\end{document}